\pdfoutput=1

\documentclass[11pt]{article}
\usepackage[table]{xcolor}
\usepackage{makecell}
\usepackage{arabtex}
\usepackage{utf8}
\setcode{utf8}
\usepackage{float}
\usepackage{makecell}
\usepackage{tabularray}

\usepackage[preprint]{ACL}
\usepackage{graphicx}
\usepackage{times}
\usepackage{latexsym}
\usepackage{booktabs}
\usepackage{multirow}
\usepackage{fontawesome5}
\usepackage{pdflscape}
\usepackage{tabularx}
\usepackage{ltablex}
\usepackage{longtable}
\usepackage{romanbar}
\usepackage{stfloats}
\usepackage{array}
\usepackage{geometry}
\usepackage{lipsum}
\usepackage{caption}
\usepackage[T1]{fontenc}

\usepackage[utf8]{inputenc}

\usepackage{microtype}

\title{WTU-EVAL: A Whether-or-Not Tool Usage Evaluation Benchmark for Large Language Models}

\author{Kangyun Ning\textsuperscript{1,\thanks{*Equal contribution}} , Yisong Su\textsuperscript{2,\footnotemark[1]}, Xueqiang Lv, Yuanzhe Zhang\textsuperscript{3,\thanks{\textsuperscript{+} Corresponding authors}}, Jian Liu\textsuperscript{1,\footnotemark[2]},Kang Liu\textsuperscript{3}, Jinan Xu\textsuperscript{1}\\
    $^{1}$Beijing Key Lab of Traffic Data Analysis and Mining,\\
    Beijing Jiaotong University, Beijing, China \\
    $^{2}$College of Computer and Data Science, Fuzhou University \\
    $^{3}$The Laboratory of Cognition and Decision Intelligence for Complex Systems,\\
    Institute of Automation, CAS\\
    \texttt{\{22120409, jianliu, jaxu\}@bjtu.edu.cn}, \texttt{221020042@fzu.edu.cn}
  }

\begin{document}
\maketitle
\begin{abstract}
Although Large Language Models (LLMs) excel in NLP tasks, they still need external tools to extend their ability. 
Current research on tool learning with LLMs often assumes mandatory tool use, which does not always align with real-world situations, where the necessity for tools is uncertain, and incorrect or unnecessary use of tools can damage the general abilities of LLMs.
Therefore, we propose to explore whether LLMs can discern their ability boundaries and use tools flexibly. We then introduce the Whether-or-not tool usage Evaluation benchmark (WTU-Eval) to assess LLMs with eleven datasets, where six of them are tool-usage datasets, and five are general datasets. LLMs are prompted to use tools according to their needs. 
The results of eight LLMs on WTU-Eval reveal that LLMs frequently struggle to determine tool use in general datasets, and LLMs' performance in tool-usage datasets improves when their ability is similar to ChatGPT. In both datasets, incorrect tool usage significantly impairs LLMs' performance. 
To mitigate this, we also develop the finetuning dataset to enhance tool decision-making.
Fine-tuning Llama2-7B results in a 14\% average performance improvement and a 16.8\% decrease in incorrect tool usage. We will release the WTU-Eval benchmark.
\end{abstract}

\section{Introduction}

\begin{figure}
  \centering
  \includegraphics[width=0.9\linewidth]{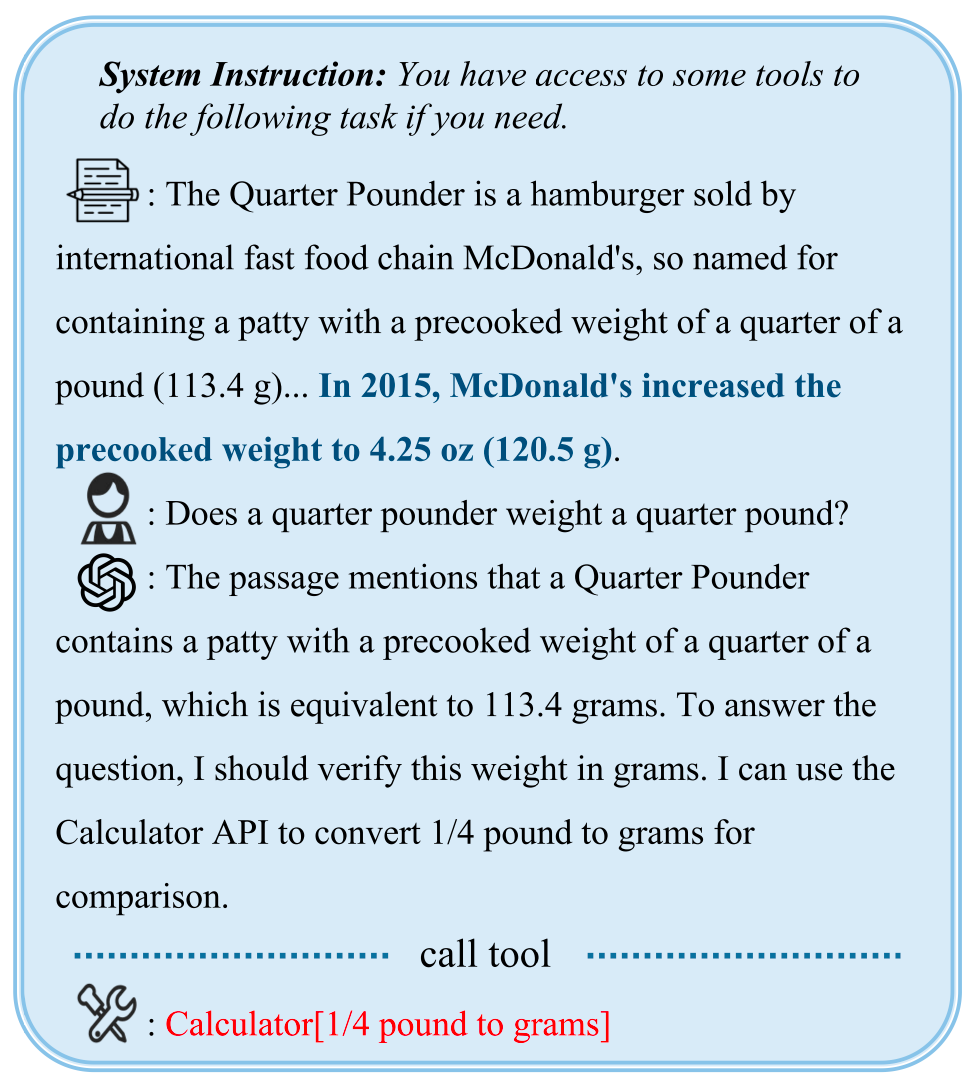}
  \caption{An example showing the failure of calling tools inappropriately.}
  \label{fig:timing}
\end{figure}

LLMs like GPT4 \cite{openai2023gpt4} excel in general-purpose NLP tasks, which however struggle in tasks requiring calling external tools, such as fetching up-to-date stock prices and booking flights \cite{qin2023tool,qin2023toolllm,patil2023gorilla,ruan2023tptu}.
How to improve LLMs' interaction with tools is a current hot topic.
Recent studies have explored tool usage fine-tuning \cite{qin2023toolllm,patil2023gorilla} and reinforcement learning \cite{li2023tool} techniques, showing promising results in areas like math reasoning and web search. 

Despite notable progress, prior studies \cite{patil2023gorilla,zhuang2023toolqa} mainly focused on scenarios mandating tool use by LLMs.
However, in a real-world application, the necessity for tool usage is uncertain. Moreover, we observe that inappropriate tool invocation can lead to errors, adversely affecting outcomes.
For example, Figure~\ref{fig:timing} provides an example of using ChatGPT (0613) to answer a question.
Despite the context hinting at the answer: a quarter pounder's weight has been increased to 120.5g, not a quarter pound (113.4g), ChatGPT still invokes an external tool, \textit{Calculator}, and due to incorrect parameter settings, it produces an erroneous response and redundant response time.

With the above observations, we want to explore an intriguing question: whether LLMs can discern their ability boundaries, and if LLMs have the option to decide whether to use tools, would their performance improve in general and tool-usage datasets?

To this end, we propose a Whether-or-not tool usage Evaluation benchmark (WTU-Eval), which contains six tool-usage datasets that explicitly require tool usage and five general datasets that can be answered without tools. 
As illustrated in the accompanying Figure~\ref{fig:figure1}, Region1 (R1) and Region3 (R3) are baselines that test LLMs without tools, and Region2 (R2) and Region4 (R4) evaluate LLMs that have the option to use tools flexibly according to their needs.

Furthermore, we also develop a dataset from the WTU-Eval benchmark training sets, resulting in a finetuning dataset with a size of 4000.
This dataset is used to enhance the model's decision-making capability regarding tool use, resulting in a 14\% average performance improvement and a 16.8\% decrease in incorrect tool usage with finetuning Llama2-7B, which also gains a significant improvement by up to 40\% for the PIQA's Search Engine—and reduces the tool invocation rate ($\S\ref{sft}$ ).

The contributions of this paper are as follows:
\begin{itemize}
    \item We propose to explore whether LLMs can discern their ability boundaries and use tools flexibly and introduce the WTU-Eval, which is the first benchmark to evaluate whether to use tools accurately.
    
    \item We rigorously evaluate the performance of eight well-known LLMs and highlight their limitations. Most LLMs struggle to recognize their capability boundaries and lack of tool usage decision-making capability.
    
    \item Based on the above insights, we also introduce a finetuning dataset, particularly for enhancing the model's decision-making capability regarding tool use, showing its positive effects.
\end{itemize}



\section{Related Work}

Integrating tool calls into LLMs spans three critical areas: API collection and search, tool assistant strategy, and performance evaluations.

\textbf{API Collection and Search.} APIBench \cite{patil2023gorilla}, featuring APIs from HuggingFace, TorchHub, and TensorHub, assesses its proficiency.
ToolBench \cite{qin2023toolllm} features 16000+ real-world APIs across 49 categories from RapidAPI Hub, and develops a depth-first search decision tree (DFSDT), improving LLMs' search and reasoning capabilities.

\textbf{Tool Assistant Strategy.} SelfAsk \cite{press2022measuring} simplifies tasks into sub-questions for tool invocation, akin to DemonstrateSearch-Predict (Khattab et al., 2023). Similarly, Toolformer \cite{schick2023toolformer}, ART \cite{paranjape2023art}, and others \cite{gao2023pal,lyu2023faithful,chen2022program} using specific tokens to guide tool usage, halting to invoke tools, and incorporating their outputs for continued generation. But they only focus on tool-usage tools, and can not apply to real-world scenarios.

\textbf{Tool Usage Evaluation.} \citet{jacovi2023comprehensive} focuses on mathematical reasoning and reveals the influence of tool use versus non-use is less pronounced in larger LLMs compared to smaller ones. MetaTool \cite{huang2023metatool} assesses LLMs' decisions on whether to utilize external tools and which tool to use, but does not address the effects of incorrect or unnecessary tool usage. 

Different from previous works, WTU-Eval aims to bridge this gap by investigating whether models recognize the need for tool use in real-world scenarios and how improper tool integration might affect the foundational efficiency of LLMs, as detailed in $\S\ref{sec:detailed-discussion}$.

\section{The WTU-Eval Benchmark}\label{sec:detailed-discussion}

\begin{figure*}[h]
\centering
\includegraphics[width=\textwidth]{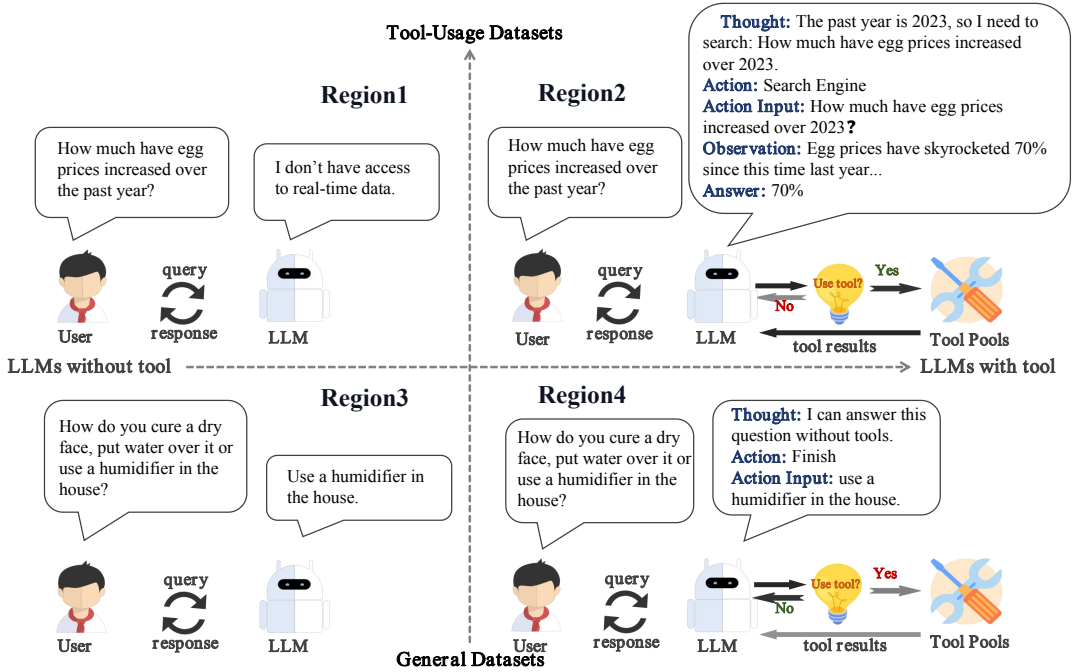}
 \caption{Illustrative diagram depicting user interaction scenarios with and without access to tool pools. LLMs need to respond to the user's query in \textit{Region1} (R1) and \textit{Region3} (R3). In \textit{Region2} (R2) and \textit{Region4} (R4), LLMs must judge based on the nature of the task whether a tool is required. If so, the corresponding tool from the tool pool is invoked; if not, the answer is provided using its knowledge. If the judgment is correct, then the corresponding choice is highlighted in \textcolor{green}{green}; otherwise, it is in \textcolor{red}{red}.}
\label{fig:figure1}
\end{figure*}

The desired diagram of WTU-Eval is shown in Figure~\ref{fig:figure1}. In R1, the user asks a real-time question, but LLM cannot access this information without the search engine, so it fails to answer. In R2, when faced with the same question, LLM has access to tool pools and knows that the tool usage is necessary, so it decides to call \textit{Search Engine} to find the real-time information and gives the correct answer. In R3, the user asks a general question, and LLM answers it with its knowledge. In R4, when presented with the same question, the LLM can access tool pools. Recognizing that tool usage is unnecessary, it decides to provide an answer directly.

By comparing the results between R1 and R2, we can determine whether LLMs recognize when a question exceeds their capabilities and thus requires the use of tools, and quantify the impact of using tools. By comparing the results of R3 and R4, we can determine whether the LLMs, when given the option to use tools, recognize that the current question can be answered without tools. Additionally, we can quantify the damage when they choose to use tools unnecessarily.

\subsection{Evaluation Settings}

We show \textbf{WTU-Eval} settings from datasets, tool pools, LLMs, and evaluation metrics.

\paragraph{Datasets.}

We partition the datasets into the tool datasets (for tasks requiring specific tools), and the general datasets (for tasks solvable with LLMs' own ability).
The tool datasets include MLQA \cite{lewis2019mlqa}, ASDiv \cite{miao2021diverse}, GSM8K \cite{cobbe2021training}, MathQA \cite{amini2019mathqa}, HotpotQA \cite{yang2018hotpotqa}, and RealtimeQA \cite{kasai2022realtime}, focusing on machine translation, math reasoning, Wikipedia search, and web search.
The general datasets contain BoolQ \cite{clark2019boolq}, RACE \cite{lai2017race}, PIQA \cite{bisk2020piqa}, RTE \cite{dagan2005pascal}, and HellaSwag \cite{zellers2019HellaSwag}, focusing on reading comprehension, commonsense reasoning, and sentence completion.
More details about the datasets are discussed in the Appendix~\ref{dataset}.

\paragraph{Tool Pools.}

Following BMTools \cite{qin2023tool}, we select the tools used in the evaluation, where machine translator and calculator are single-action tools, and search engine and Wikipedia search are multiple-action tools.

\begin{itemize}
    \item\textbf{Machine Translator}: We select Baidu Translator\footnote{\url{https://fanyi-api.baidu.com/?fr=pcHeader}}, as a current mainstream translation API with good performance, for testing.
    \item\textbf{Calculator}:  We choose the WolframAlpha API\footnote{\url{https://developer.wolframalpha.com/}} as our calculator.
    \item\textbf{Search Engine}: We choose the Bing Search\footnote{\url{https://www.microsoft.com/en-us/bing/apis/bing-web-search-api}} API as the web search tool for LLMs to browse current events, fiction stories, history facts, etc. 
    \item\textbf{Wikipedia Search}: Besides a simple Wikipedia API, as WikiSearch and WikiLoadPage are designed, we define an additional action – WikiDisambiguation. When the search entity cannot return the expected result, the model can access the interface to get a similar entity to the current search result and self-correct the search parameters.
\end{itemize}

\paragraph{LLMs.}

We test LLMs from both commercial and open-source sectors for a broad evaluation, including Text-Davinci-003, ChatGPT (0613), Llama2, ChatGLM3-6B, and Zephyr-7B. ChatGLM3-6B is notable for its unique agent-tuning with tool interaction insights. Zephyr-7B, evolved from Mistral-7B, employs Direct Distilled Preference Optimization (DPO) to better align with user preferences in language tasks.

\paragraph{Evaluation Metrics.} In WTU-Eval, we prioritize accuracy using advanced methods beyond exact matches, categorizing datasets into numerical and free-text responses. We check numerical answers with specific data and transform free-text responses into labels. For example, in PIQA (which provides two solutions for a given task), we label these two solutions as 1 and 2. When we cannot match labels or text content, we manually check the responses. \footnote{If both solutions are deemed unsuitable: \texttt{"answer": "Neither solution is suitable"}\newline If the model discusses both solutions: \texttt{"answer": "Solution 1 is..., solution 2 is...., I think solution 1 is better"}}

Additionally, tool usage is marked incorrect in the general dataset and a correct example is shown in Figure~\ref{fig:figure1} R4. To balance comparisons, we introduce the Call Rate, considering the initial use of a tool as a call, ensuring a thorough evaluation.

\subsection{Evaluation Prompt}

In WTU-Eval, we utilize ReACT \cite{yao2022react} for zero-shot and few-shot experiments in scenarios with access to tool pools (R2 and R4). The ReACT is structured into four stages: \textit{Thought}, \textit{Action}, \textit{Observation}, and \textit{Final Answer}, performed in a limited loop. To ensure fairness, all LLMs are evaluated under the same settings during the assessment. Further details on prompts are provided in the Appendix~\ref{prompts}.

\paragraph{Zero-shots.}

We introduce tool names, descriptions, and parameters to guide the LLMs to use tools.

\paragraph{Few-shots.}

We introduce tool names, descriptions, parameters, and examples of: a) one tool usage scenario, and b) one general scenario where tools are not used.

\section{Experiments and Results}

\begin{table*}[ht]
\small
\centering
\begin{tabular}{l|rrrrrr|rrrrrr}
\toprule
\multirow{3}{*}{\textbf{Test Set}} & \multicolumn{6}{c}{\textbf{Model w/o Tool}} &\multicolumn{6}{c}{\textbf{Model w/ Tool}}\\
 & \multirow{2}{*}{T003} & \multirow{2}{*}{ChatGPT} & \multicolumn{2}{c}{Llama2-13B} & \multicolumn{2}{c}{Llama2-7B} & \multirow{2}{*}{T003} & \multirow{2}{*}{ChatGPT} & \multicolumn{2}{c}{Llama2-13B} & \multicolumn{2}{c}{Llama2-7B}\\
  & & &Base&Chat&Base&Chat&&&Base&Chat&Base&Chat\\
 \midrule
 \multicolumn{2}{l}{\textbf{Tool datasets}}&\multicolumn{5}{c}{\textbf{R1}}&\multicolumn{6}{c}{\textbf{R2}}\\
 \midrule
\multirow{2}{*}{MLQA}&\multirow{2}{*}{54.17} & \multirow{2}{*}{53.13} & \multirow{2}{*}{\textbf{52.08}} & \multirow{2}{*}{57.29} & \multirow{2}{*}{\textbf{55.21}} & \multirow{2}{*}{\textbf{62.50}} & 58.33 & 50.00 &0.00 & 12.50 & 1.04 & 11.45\\
 & & & & & & & \cellcolor{yellow!50}\textbf{70.83} & \cellcolor{yellow!50}\textbf{65.62} & \cellcolor{yellow!50}26.04 & \cellcolor{yellow!50}\textbf{60.41} & \cellcolor{yellow!50}50.00 & \cellcolor{yellow!50}48.95\\
 \midrule
 \multirow{2}{*}{ASDiv}&\multirow{2}{*}{48.67} & \multirow{2}{*}{79.33} & \multirow{2}{*}{\textbf{13.00}} & \multirow{2}{*}{\textbf{50.00}} & \multirow{2}{*}{23.00} & \multirow{2}{*}{45.67} & \textbf{70.66} & \textbf{83.00} &9.00 & 23.66 & \textbf{46.66} & 38.66\\
 & & & & & & & \cellcolor{yellow!50}68.33 & \cellcolor{yellow!50}\textbf{83.00} & \cellcolor{yellow!50}9.00 & \cellcolor{yellow!50}45.00 & \cellcolor{yellow!50}43.00 & \cellcolor{yellow!50}\textbf{47.66}\\
\midrule
 \multirow{2}{*}{GSM8K}&\multirow{2}{*}{14.00} & \multirow{2}{*}{67.00} & \multirow{2}{*}{\textbf{2.00}} & \multirow{2}{*}{9.00} & \multirow{2}{*}{9.00} & \multirow{2}{*}{\textbf{12.00}} & 39.00 & \textbf{58.00} &0.00 & \textbf{20.00}& 2.00 & 8.00\\
 & & & & & & & \cellcolor{yellow!50}\textbf{52.00} & \cellcolor{yellow!50}53.00 & \cellcolor{yellow!50}0.00 & \cellcolor{yellow!50}15.00 & \cellcolor{yellow!50}\textbf{14.00} & \cellcolor{yellow!50}5.00\\
\midrule
 \multirow{2}{*}{MathQA}&\multirow{2}{*}{33.00} & \multirow{2}{*}{18.00} & \multirow{2}{*}{\textbf{12.00}} & \multirow{2}{*}{\textbf{17.00}} & \multirow{2}{*}{\textbf{19.00}} & \multirow{2}{*}{\textbf{26.00}} & 37.00 & \textbf{39.00} &8.00 & 11.00 & 4.00 & 11.00\\
 & & & & & & & \cellcolor{yellow!50}\textbf{39.00} & \cellcolor{yellow!50}36.00 & \cellcolor{yellow!50}6.00 & \cellcolor{yellow!50}12.00 & \cellcolor{yellow!50}10.00 & \cellcolor{yellow!50}5.00\\
\midrule
 \multirow{2}{*}{RealtimeQA}&\multirow{2}{*}{36.66} & \multirow{2}{*}{\textbf{40.00}} & \multirow{2}{*}{\textbf{20.00}} & \multirow{2}{*}{30.00} & \multirow{2}{*}{23.34} & \multirow{2}{*}{40.00} & \textbf{56.66} & 36.66 &0.00 & 23.33 & 3.30 & 40.00\\
 & & & & & & & \cellcolor{yellow!50}36.66 & \cellcolor{yellow!50}40.00 & \cellcolor{yellow!50}0.00 & \cellcolor{yellow!50}\textbf{33.33} & \cellcolor{yellow!50}\textbf{26.66} & \cellcolor{yellow!50}\textbf{26.66}\\
 \midrule
 \multirow{2}{*}{HotPotQA}&\multirow{2}{*}{33.50} & \multirow{2}{*}{34.50} & \multirow{2}{*}{\textbf{11.50}} & \multirow{2}{*}{\textbf{33.00}} & \multirow{2}{*}{\textbf{20.00}} & \multirow{2}{*}{\textbf{36.00}} & 28.50 & 39.00 &0.00 & 18.00 & 0.00 & 20.50\\
 & & & & & & & \cellcolor{yellow!50}\textbf{47.95} & \cellcolor{yellow!50}\textbf{41.50} & \cellcolor{yellow!50}5.00 & \cellcolor{yellow!50}26.50 & \cellcolor{yellow!50}18.50 & \cellcolor{yellow!50}20.50\\
 \midrule
\multicolumn{2}{l}{\textbf{General datasets}}&\multicolumn{5}{c}{\textbf{R3}}&\multicolumn{6}{c}{\textbf{R4}}\\
 \midrule
\multirow{2}{*}{BoolQ}&\multirow{2}{*}{\textbf{79.00}} & \multirow{2}{*}{\textbf{89.00}} & \multirow{2}{*}{\textbf{56.00}} & \multirow{2}{*}{46.00} & \multirow{2}{*}{46.00} & \multirow{2}{*}{\textbf{57.00}} & 20.00 & 6.00 &0.00 & 0.00 & 0.00 & 2.00\\
 & & & & & & & \cellcolor{yellow!50}58.50 & \cellcolor{yellow!50}76.25 & \cellcolor{yellow!50}54.00 & \cellcolor{yellow!50}\textbf{61.25} & \cellcolor{yellow!50}\textbf{55.00} & \cellcolor{yellow!50}32.50\\
 \midrule
 \multirow{2}{*}{RACE}&\multirow{2}{*}{68.96} & \multirow{2}{*}{\textbf{79.09}} & \multirow{2}{*}{14.80} & \multirow{2}{*}{22.93} & \multirow{2}{*}{32.00} & \multirow{2}{*}{33.87} & 6.00 & 30.00 &0.00 & 0.00 & 14.00 & 0.00\\
 & & & & & & & \cellcolor{yellow!50}\textbf{82.93} & \cellcolor{yellow!50}77.46 & \cellcolor{yellow!50}\textbf{62.05} & \cellcolor{yellow!50}\textbf{52.53} & \cellcolor{yellow!50}\textbf{58.40} & \cellcolor{yellow!50}\textbf{50.40}\\
\midrule
 \multirow{2}{*}{PIQA}&\multirow{2}{*}{\textbf{58.00}} & \multirow{2}{*}{\textbf{84.00}} & \multirow{2}{*}{16.00} & \multirow{2}{*}{32.00} & \multirow{2}{*}{\textbf{25.00}} & \multirow{2}{*}{\textbf{49.00}} & 3.00 & 39.00 &0.00 & 0.00 & 0.00 & 0.00\\
 & & & & & & & \cellcolor{yellow!50}50.25 & \cellcolor{yellow!50}58.75 & \cellcolor{yellow!50}\textbf{27.25} & \cellcolor{yellow!50}\textbf{43.50} & \cellcolor{yellow!50}18.75 & \cellcolor{yellow!50}31.00\\
\midrule
 \multirow{2}{*}{RTE}&\multirow{2}{*}{59.00} & \multirow{2}{*}{\textbf{78.00}} & \multirow{2}{*}{\textbf{68.00}} & \multirow{2}{*}{\textbf{63.00}} & \multirow{2}{*}{\textbf{54.00}} & \multirow{2}{*}{\textbf{58.00}} & 3.00 & 12.00 &0.00 & 0.00 & 0.00 & 0.00\\
 & & & & & & & \cellcolor{yellow!50}\textbf{66.25} & \cellcolor{yellow!50}50.00 & \cellcolor{yellow!50}13.50 & \cellcolor{yellow!50}34.50 & \cellcolor{yellow!50}45.50 & \cellcolor{yellow!50}23.25\\
\midrule
 \multirow{2}{*}{HellaSwag}&\multirow{2}{*}{\textbf{54.00}} & \multirow{2}{*}{\textbf{75.00}} & \multirow{2}{*}{\textbf{21.00}} & \multirow{2}{*}{\textbf{49.00}} & \multirow{2}{*}{\textbf{54.00}} & \multirow{2}{*}{\textbf{44.00}} & 23.00 & 28.00 &0.00 & 0.00 & 1.00 & 0.00\\
 & & & & & & & \cellcolor{yellow!50}50.75 & \cellcolor{yellow!50}50.00 & \cellcolor{yellow!50}4.25 & \cellcolor{yellow!50}23.25 & \cellcolor{yellow!50}20.00 & \cellcolor{yellow!50}15.75\\

\bottomrule
\end{tabular}
\caption{The accuracy of experiments executed general datasets and tool datasets whether or not have access to tool pools, where "T003" means "Text-Davinci-003", and "\colorbox{yellow!50}{\quad}" indicates few-shot results, while cells without background color indicate zero-shot results.}
\label{tab:RQ1}
\end{table*}

\subsection{LLMs' Performance in Tool Datasets}

\textbf{When LLMs can determine whether to use tools and LLMs' ability is similar to ChatGPT, their performance in tool-usage datasets improves.}
In Table~\ref{tab:RQ1} R2, when LLMs have access to tools, Llama2-13B's zero-shot performance on most tool questions drops to 0, while ChatGPT and Text-Davinci-003 exhibit significant improvements (by up to 25\% in GSM8K), exceeding their performance in R1. It is observed that the use of tools does not unconditionally enhance LLMs' performance and the enhancement depends on LLMs' ability. Considering the scale gap between ChatGPT, Text-Davinci-003, and Llama2, we believe that properly using tools demands models' ability to deal with complex and extensive tool prompts without demonstrations.

This trend alters a little with the adoption of the few-shot methodology. 
In R2, ChatGPT and Text-Davinci-003's performance also improve (by up to 40\% in GSM8K) with the few-shot setting, exceeding their performance in R1. In contrast, Llama2 only shows improvement on a small portion of tool datasets, with performance declining on the rest compared to R1.
It is concluded that the efficacy of tool invocation in augmenting performance is contingent upon the ability of the model.

\subsection{Impact of Different Tools on LLMs' Performance in Tool Datasets}

\textbf{In most tool-usage datasets, the proficiency of LLMs diminishes as the complexity of tools increases.}
In Table~\ref{tab:RQ1} R2, we especially introduce the Translator to MLQA, the Calculator to ASDiv, GSM8K, and MathQA, the Search Engine to RealtimeQA, and the Wikipedia Search to HotPotQA.

Tool usage impact is closely linked to tool complexity. LLMs efficiently manage translation tasks due to the Translator's simplicity. However, when faced with complex tools like the WolframAlpha Calculator, Llama2's performance drops significantly. Similarly, tasks using BingSearch and WikipediaSearch see only modest improvements due to more complex tool instructions, particularly in the few-shot setting across all LLMs.

Moreover, the few-shot setting remarkably outperforms the zero-shot setting, with improvements reaching up to 76\% in some cases. In zero-shot settings, such as ChatGPT's use of a range of tools from Translator to Wikipedia Search, there is a clear trend: as tool's complexity increases, LLMs' proficiency decreases. This indicates that tasks requiring a deeper understanding of tool usage present more significant challenges for LLMs, underscoring the increased interpretive burden in navigating tool-specific instructions.

\subsection{LLMs' Performance in General Datasets}

\textbf{LLMs' performance in general datasets declines when they can determine whether to use tools, indicating LLMs do not know their ability boundary.}
By comparing R4 to R3 in Table~\ref{tab:RQ1}, we can observe that LLMs' performance decreases in all general datasets. Analyzing the incorrect answer, we note that LLMs tend to use tools, and due to wrong tool invocation, their performance declines. The whole incorrect answer study will be discussed in $\S\ref{casestudy}$.

Table~\ref{tab:RQ1} R4 demonstrates a significant reduction in zero-shot performance when accessing tools compared with R3, particularly evident in Llama2. Notably, the most substantial decrease observed is 83\% in BoolQ (Text-Davinci-003), and Llama2's performance nearly falls to 0. This is primarily due to LLMs' frequent misuse of tools in general queries. Error analysis \$\ref{casestudy} suggests that the complexity of following tool instructions complicates the adherence to the ReACT framework, thus impacting the \textit{Thought} process.

To mitigate this issue, we add demonstrations, leading to the few-shot results in R4. These experiments largely echo the zero-shot findings, but slight improvements are observed compared with R3. Importantly, this increase is mainly observed in Llama2, which shows a 10\% to 30\% improvement. We believe the demonstrations not only inspire LLMs' Chain of Thought (COT) ability but also correct their response formats.

\begin{figure}[ht]
\centering
\includegraphics[width=\columnwidth]{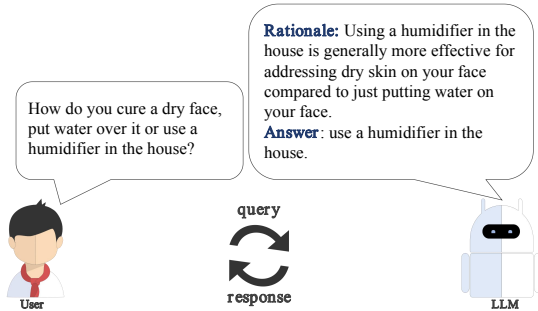}
\caption{Illustrative diagram depicting user interaction scenarios with LLMs in COT setting without the integration of a tool set.}
\label{fig:figure2}
\end{figure}

To further explore the impact of the COT and ReACT's format, we conduct few-shot trials in R3, focusing on the COT process depicted in Figure~\ref{fig:figure2}. The results, as shown in Table~\ref{tab:COT}, reveal that COT significantly aids smaller-scale Llama models. However, for larger models such as ChatGPT, COT does not lead to improvements and might even result in performance declines on BoolQ and HellaSwag.

\begin{table}[ht]
\resizebox{1.0\linewidth}{!}{
\centering
\begin{tabular}{lrrrrrr}
\toprule
\multirow{2}{*}{\textbf{Test set}} & \multicolumn{2}{c}{\textbf{ChatGPT}} & \multicolumn{2}{c}{\textbf{Llama2-7B}} & \multicolumn{2}{c}{\textbf{Llama2-7B-Chat}} \\
 & \textbf{Zero-shot} & \textbf{COT} & \textbf{Zero-shot} & \textbf{COT} & \textbf{Zero-shot} & \textbf{COT} \\
\midrule
BoolQ & \textbf{89.00} & 81.00 & 46.00 & \textbf{74.00} & \textbf{57.00} & 51.00\\
RACE & 79.09 & \textbf{83.47} & 60.30 & \textbf{65.07} & 33.87 & \textbf{67.73}\\
PIQA & 84.00 & \textbf{86.00} & 25.00 & \textbf{56.00} & 49.00 & \textbf{54.00}\\
RTE & 78.00 & 78.00 & 54.00 & \textbf{57.00} & \textbf{58.00} & 47.00\\
HellaSwag & \textbf{75.00} & 66.00 & \textbf{54.00} & 24.00 & \textbf{44.00} & 39.00\\
\bottomrule
\end{tabular}
}
\caption{Accuracy in general datasets without tool access in COT and zero-shot settings.}
\label{tab:COT}
\end{table}

\subsection{Impact of Different Tools on LLMs' Performance in General Datasets}

\begin{table}[ht]
\centering
\small
\resizebox{1.0\linewidth}{!}{
\renewcommand{\arraystretch}{1.2}
\begin{tabular}{lcrrrr}
\toprule
\textbf{LLM} & \textbf{MT} & \textbf{Cal} & \textbf{SE} & \textbf{Wiki} & \textbf{All} \\
\midrule
\multirow{2}{*}{T003 }& 56.00 & 47.00 & 31.00& \textbf{42.00} & 20.00 \\
& \cellcolor{yellow!50}\textbf{68.00}& \cellcolor{yellow!50}\textbf{83.00} & \cellcolor{yellow!50}\textbf{45.00} & \cellcolor{yellow!50}38.00 & \cellcolor{yellow!50}\textbf{58.50}\\
 \midrule
\multirow{2}{*}{ChatGPT} & 11.00 & 5.00 & 8.00 & 8.00 & 6.00 \\
& \cellcolor{yellow!50}\textbf{80.00} & \cellcolor{yellow!50}\textbf{85.00} & \cellcolor{yellow!50}\textbf{70.00} & \cellcolor{yellow!50}\textbf{70.00} & \cellcolor{yellow!50}\textbf{76.25} \\
 \midrule
\multirow{2}{*}{Llama2-7B-Base} & 0.00 & 0.00 & 0.00 & 0.00 & 0.00\\
& \cellcolor{yellow!50}\textbf{64.00} & \cellcolor{yellow!50}\textbf{59.00} & \cellcolor{yellow!50}\textbf{41.00} & \cellcolor{yellow!50}\textbf{56.00} & \cellcolor{yellow!50}\textbf{54.00}\\
 \midrule
\multirow{2}{*}{Llama2-7B-Chat}  & 0.00& 0.00 & 0.00 & 0.00 & 0.00\\
& \cellcolor{yellow!50}\textbf{45.00} & \cellcolor{yellow!50}\textbf{42.00} & \cellcolor{yellow!50}\textbf{9.00} & \cellcolor{yellow!50}\textbf{34.00} & \cellcolor{yellow!50}\textbf{32.50}\\
 \midrule
\multirow{2}{*}{Llama2-13B-Base} & 0.00 & 0.00 & 0.00 & 0.00 & 0.00\\
& \cellcolor{yellow!50}\textbf{62.00} & \cellcolor{yellow!50}\textbf{56.00} & \cellcolor{yellow!50}\textbf{46.00} & \cellcolor{yellow!50}\textbf{52.00} & \cellcolor{yellow!50}\textbf{55.00}\\
 \midrule
\multirow{2}{*}{Llama2-13B-Chat}  & 0.00 & 0.00 & 1.00 & 0.00 & 0.00\\
& \cellcolor{yellow!50}\textbf{77.00} & \cellcolor{yellow!50}\textbf{52.00} & \cellcolor{yellow!50}\textbf{60.00} & \cellcolor{yellow!50}\textbf{56.00} & \cellcolor{yellow!50}\textbf{32.50}\\
 \midrule
 \multirow{2}{*}{Zephyr-7B}  & 35.00 & \textbf{33.00} & 35.00 & 34.00 & 17.00\\
 & \cellcolor{yellow!50}\textbf{52.00} & \cellcolor{yellow!50}8.00 & \cellcolor{yellow!50}\textbf{53.00} & \cellcolor{yellow!50}\textbf{77.00} & \cellcolor{yellow!50}\textbf{47.50}\\
 \midrule
\multirow{2}{*}{ChatGLM3-6B} & 10.00 & 7.00 & 8.00 & \textbf{18.00} & 20.00\\
& \cellcolor{yellow!50}\textbf{31.00} & \cellcolor{yellow!50}\textbf{43.00} & \cellcolor{yellow!50}\textbf{23.00} & \cellcolor{yellow!50}14.00 & \cellcolor{yellow!50}\textbf{27.75}\\
\bottomrule
\end{tabular}
}
\caption{Detailed Results of BoolQ Experiment: Performance of each LLM in few-shot and zero-shot settings, where \textit{MT} means \textit{Machine Translator}, \textit{Cal} means \textit{Calculator}, \textit{SE} means \textit{Search Engine}, \textit{Wiki} means \textit{WikiPedia Search}, and "\colorbox{yellow!50}{\quad}" indicates few-shot results, while cells without background color indicate zero-shot results.}
\label{tab:BoolQ}
\end{table}

\textbf{In general datasets, LLMs experience a decline in performance when various external tools are introduced, with the impact order being (Wikipedia Search, Search Engine) more significant than (Translator, Calculator).} We test LLMs by introducing all tools and tool usage instructions. In the zero-shot setting, the collective impact of the tool pool is evaluated by introducing information for all tools simultaneously in the instructions. In the few-shot setting, due to the input length constraints of LLMs, we present each tool's name, description, and demonstration in individual prompts. The average accuracy across all tools reflects the cumulative effect of the tool pools. 
\label{Boolq}

As illustrated in Table~\ref{tab:BoolQ}, it is observed that due to the increased complexity introduced by the addition of tool instruction, zero-shot performance is considerably lower than few-shot, and reveals that different tool demonstrations have varying effects on the model's performance in general datasets, with the order of impact being (Wikipedia Search, Search Engine) more pronounced than (Translator, Calculator). The complete experiment of all general datasets is shown in Appendix~\ref{all_fs_zs_general}.

\begin{table*}[ht]
\small
\centering
\begin{tabular}{|m{2cm}|m{9cm}|m{3cm}|}
\hline
\textbf{Error Type} & \textbf{Example} & \textbf{Analysis} \\ 
\hline
 Incorrect or Unnecessary Tool Invocation & \textbf{Passage}: ...People often get take-away meals--they buy the food outside\newline
    \textbf{Question}: When they get a takeaway meal, they often eat it ()?\newline
    \textbf{Options}: A. at home B. in the school C. outside D. in the bars\newline
    \textit{\textbf{Thought:}} There is no answer in the passage.
    \textit{\textbf{Action:}} BingSearch\newline
    \textit{\textbf{Action Input:}} Where do people often eat takeaway meals? & The LLMs' judgment on when to invoke tools is inaccurate, often opting for tool usage in the first \textit{Thought} step. \\ 
\hline
 NULL Content & I have 3 hundreds, 8 tens, and 3 ones. What number am I? \newline
\textit{\textbf{Thought:}} I should try to find a way to convert the numbers into a number that I can add.
\textit{\textbf{Action:}}  & \textit{Action} is null. Insufficient training in framework comprehension leads to failures in producing the required content. \\ 
\hline
Call Correct Tool but Not Reasoning &   Jerry was weighing boxes for moving. The first was 2 pounds, the second was 11 pounds and the last was 5 pounds. What is the combined weight of all three boxes? \newline
\textit{\textbf{Thought:}} I need to calculate the combined weight of the boxes. 
\textit{\textbf{Action:}} Calculator 
\textit{\textbf{Action Input:}} 2 + 11 + 5 
\textit{\textbf{Observation:}}  \{`@title': `Result', `@primary': `true', `subpod': \{`@title': `', `img': \{`@src':\{given link\}, `@alt': `18', `@title': `18', `@contenttype': `image/gif'\}\newline
\textit{\textbf{Thought:}} I need to calculate the combined weight of the boxes. & The model pays more attention to tool usage rather than understanding the returned observation and moving to the next state. \\ 
\hline
 Call Invalid Tool Repeatedly After Reminding & 14 red plums and four green plums are in the basket. How many plums are in the basket?\newline
\textit{\textbf{Thought:}} I should use a calculator for Addition to finish this task.
\textit{\textbf{Action:}} Calculator
\textit{\textbf{Action Input:}} 14 (red plums) + 4 (green plums)
\textit{\textbf{Observation:}} WolframAlpha API cannot parse the input query.
\textit{\textbf{Thought:}} Hmm, I think I need to separate the red and green plums.
\textit{\textbf{Action:}} Calculator
\textit{\textbf{Action Input:}} 14 (red plums) + 4 (green plums) & The model cannot comprehend the instruction and forgets how to consistently use the given tools. \\ 
\hline
Endless Retry & The McCarran Airport Connector serves what airport named after late U.S. Senator Pat McCarran?\newline
    \textit{\textbf{Thought:}} I need to find information about the McCarran Airport Connector and the airport it serves, which is named after a late U.S. Senator.
    \textit{\textbf{Action:}} WikiSearch
    \textit{Action Input:} McCarran Airport Connector\newline
    \textit{\textbf{Observation:}} State Route 171. The Harry Reid Airport Connector is ...\newline
    \textit{\textbf{Thought:}} I have found information about the McCarran Airport Connector, but I need to find information about the airport it serves, which is named after a late U.S. Senator.
    \textit{\textbf{Action:}} WikiSearch
    \textit{\textbf{Action Input:}} Pat McCarran
    \textit{\textbf{Observation:}} Patrick Anthony McCarran (August 8, 1876 – September 28, 1954) ...\newline
    \textit{\textbf{Thought:}} ...& 
   The model keeps using tools in a loop without reaching a conclusion and needs to improve its ability to give up and restart when stuck in a loop. \\ 
\hline
\end{tabular}
\caption{Examples and analysis of five error types encountered in the failure cases.}
\label{tab:failure_analysis}
\end{table*}

\section{Discussion}

For a deeper understanding of the results, we explore different finetuning methods' impacts, conduct the error analysis, and make some improvements on WTU-Eval.

\subsection{Quantitative Results on Different Fine-tuning Methods}

We conduct R2 and R4's evaluation in ChatGLM3-6B and Zephyr-7B for their different fine-tuning methods from Llama2.

\paragraph{Poor Performance of ChatGLM3-6B in Decision on Whether to Use the Tool.}

 The above analysis reveals that LLMs' performance drops in general datasets with tool access but improves in tool usage datasets, dependent on LLM ability. LLMs lacking effective tool-usage training show weaker decision-making on tool employment. Thus, we experiment with ChatGLM3-6B in general and tool-usage datasets, leveraging its agent tuning for better tool usage.

\begin{table}[t]
\small
\centering
\resizebox{1.0\linewidth}{!}{
\begin{tabular}{lrrrr}
\toprule
  \multirow{2}{*}{\textbf{Test Set}} & \multicolumn{2}{c}{\textbf{ChatGLM3-6B}} & \multicolumn{2}{c}{\textbf{Llama2-7B}}\\
  & \textbf{Zero-Shot}&\textbf{Few-Shot}&\textbf{Zero-Shot}&\textbf{Few-Shot}\\
  \midrule
  \multicolumn{5}{l}{\textbf{Tool Datasets}}\\
  \midrule
  MLQA & 14.58 & 42.70& 1.04 & \textbf{50.00}\\
  ASDiv & 38.33 &\textbf{ 52.66 }& 43.66& 43.00\\
  GSM8K & 14.00 & \textbf{26.00}&2.00&14.00 \\
  MathQA & \textbf{13.00} & 9.00 &4.00&10.00\\
  RealtimeQA & 13.33 & 23.33&3.30&\textbf{26.66} \\
  HotPotQA & 4.00 & 11.50 &0.00&\textbf{18.50} \\
\midrule
\multicolumn{5}{l}{\textbf{General Datasets}}\\
  \midrule
  BoolQ & 20.00 &27.75&0.00&\textbf{55.00} \\
  RACE & 22.00 & \textbf{66.06}&14.00&58.40 \\
  PIQA & 3.00 & \textbf{19.00}&0.00&18.75 \\
  RTE & 35.00 & \textbf{46.25}&0.00&45.00 \\
  HellaSwag & 11.00 & \textbf{23.25}&1.00&20.00 \\
  \bottomrule
\end{tabular}
}
\caption{Accuracy of ChatGLM3-6B with access to tools in tool-usage datasets and general datasets.}
\label{tab:chatglm3}
\end{table}

As illustrated in Table~\ref{tab:chatglm3}, compared with the LLama2-7B, ChatGLM3-6B shows superior performance in the zero-shot settings, indicating the validity of agent tuning. However, the results also show its poor performance in the decision on whether to use the tool for similar results to Llama2-7B in most general datasets, which indicates that current tool training methods have not adequately addressed the question of whether to invoke a tool.

\begin{table}[ht]
\small
\centering
\begin{tabular}{lrrr}
\toprule
\multirow{2}{*}{\textbf{Test set}} & \textbf{Model w/o Tool} & \multicolumn{2}{c}{\textbf{Model w/ Tool}} \\
 & \textbf{Zero-Shot} & \textbf{Zero-Shot} & \textbf{Few-Shot} \\
\midrule
MLQA & \textbf{46.15} & 39.58 & 33.33 \\
AsDiv & \textbf{64.00} & 46.33 & 0.00 \\
GSM8K & 18.00 & \textbf{35.00} & 7.00 \\
MathQA & \textbf{21.00} & 7.00 & 5.00 \\
RealtimeQA & \textbf{50.00} & 23.33 & 10.00 \\
HotpotQA & \textbf{30.00} & 18.00 & 16.23 \\
\midrule
BoolQ & \textbf{76.00} & 17.00 & 47.50 \\
RACE & \textbf{75.46} & 17.00 & 59.40 \\
PIQA & \textbf{68.00} & 4.00 & 15.75 \\
RTE & \textbf{66.00} & 1.00 & 15.50 \\
HellaSwag & \textbf{31.00} & 4.00 & 15.50\\
\bottomrule
\end{tabular}
\caption{Accuracy of Zephyr-7B in tool usage and general datasets with and without tool access.}
\label{tab:zephyr}
\end{table}

\paragraph{Counterintutive Results in Zephyr-7B.} In our study, Zephyr-7B exhibits unique performance trends compared to other LLMs, particularly underlined by a decrease in efficacy when using tools in tool datasets, as detailed in Table~\ref{tab:zephyr}. Its few-shot performance falls short of its zero-shot capabilities, a discrepancy most evident in calculator-involved tasks, dropping to as low as 0 in ASDiv.
Moreover, within general datasets, the calculator's negative impact on Zephyr-7B is notably worse than that of other tools ($\S\ref{Boolq}$ ), resulting in a 38.5\% lower average accuracy. Analysis of errors in these datasets reveals a recurrent issue: Zephyr-7B frequently misapplies the calculator in calculation-related tasks, leading to response inaccuracies.

\begin{table}[ht]
    \centering
    \resizebox{1.0\linewidth}{!}{
    \begin{tabular}{lrrrrrr}
    \toprule
       \textbf{Test Set} & \textbf{Model} & \textbf{MT} & \textbf{Cal} & \textbf{SE} & \textbf{Wiki} & \textbf{All}\\
        \midrule
        \multirow{4}{*}{BoolQ} & Baseline(Acc) & 45.00 & 42.00 & 9.00 & 34.00 & 32.50\\
         & Ours(Acc) & \textbf{45.00} & \textbf{53.00} & \textbf{44.00} & \textbf{37.00} & \textbf{44.75}\\
         & Baseline(CR) & 0.00 & 4.00 & 41.00 & 10.00 & 13.75\\
         & Ours(CR) & \textbf{0.00} & \textbf{1.00} & \textbf{3.00} & \textbf{1.00} & \textbf{1.25}\\
         \midrule
         \multirow{4}{*}{PIQA} & Baseline(Acc) & \textbf{51.00} & 14.00 & 0.00 & 10.00 & 18.75\\
         & Ours(Acc) & 37.00 & \textbf{47.00} & \textbf{40.00} & \textbf{43.00} & \textbf{41.75}\\
         & Baseline(CR) & 2.00 & 80.00 & 53.00 & 11.00 & 36.50\\
         & Ours(CR) & \textbf{0.00} & \textbf{6.00} & \textbf{2.00} & \textbf{1.00} & \textbf{2.25}\\
         \midrule
         \multirow{4}{*}{HellaSwag} & Baseline(Acc) & 19.00 & 21.00 & 21.00 & 19.00 & 20.00\\
         & Ours(Acc) & \textbf{25.00} & \textbf{28.00} & \textbf{30.00} & \textbf{25.00} & \textbf{27.00}\\
         & Baseline(CR) & 0.00 & \textbf{0.00} & 6.00 & 14.00 & 5.00\\
         & Ours(CR) & \textbf{0.00} & 4.00 & \textbf{0.00} & \textbf{1.00} & \textbf{1.25}\\
         \bottomrule
    \end{tabular}
    }
    \caption{Performance on general datasets improves through SFT of Llama2-7B model, where \textit{CR} means \textit{Call Rate}.}
    \label{tab:finetuning}
\end{table}

\subsection{Error Analysis}
\label{casestudy}

In this section, we make a deep sampling of the failure cases. 
Besides the wrong answer, we set five error types of these cases, show examples and analysis in Table~\ref{tab:failure_analysis}, and more details in Appendix \ref{appendix:ErrorType}.

\begin{figure}[ht]
    \centering
    \includegraphics[width=\columnwidth]{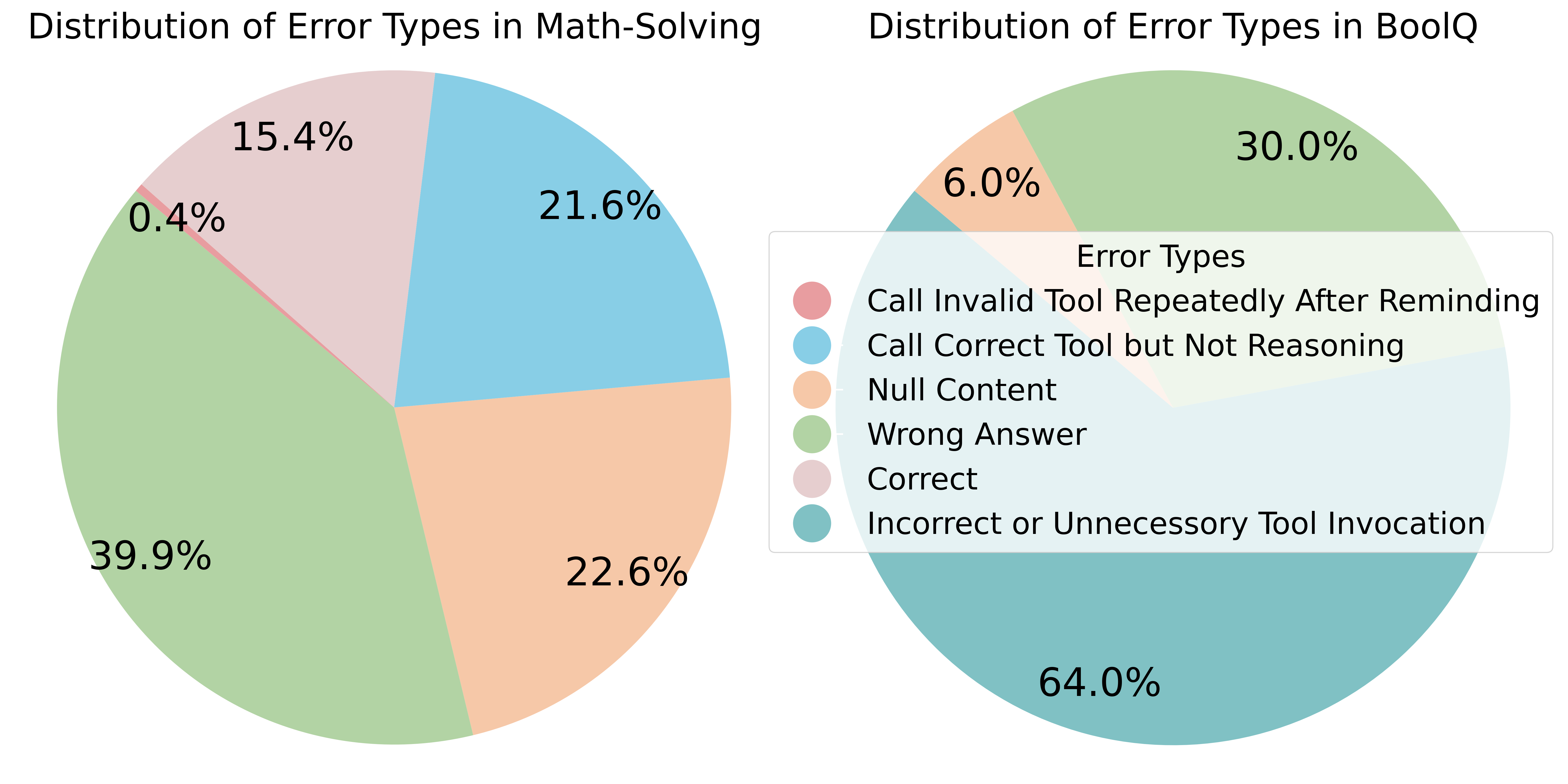}
    \caption{Distribuion of Error Types in Tool-Usage and General Datasets with Zero-Shot Setting in Llama2-7B}
    \label{fig:case-label}
\end{figure}

The proportions of these error types vary across different models, datasets, and settings. For instance, Figure~\ref{fig:case-label} shows the distribution of error types in math-solving questions (ASDiV, GSM8K and MathQA) and commonsense reason questions (BoolQ) with Llama2-7B in R2 and R4. It can be inferred that incorrect/unnecessary tool invocation is preferred to appear in general datasets, while the other error types about tool invocation steps appear in tool-usage datasets.

\subsection{Supervised Fine-Tuning for Tool-Usage Decision-Making}
\label{sft}

Based on our findings, LLMs' indecision on tool usage not only undermines their overall performance but also adversely affects their effectiveness on general datasets. To mitigate this, we curate a specialized dataset with a size of 4000 from the general datasets' training sets. Based on observation of step \textit{Thought}'s importance for the decision on tool usage in Table~\ref{tab:failure_analysis}, we train the first \textit{Thought} and second \textit{Action} steps, aiming at improving decision-making ability regarding tool usage. We apply GPT-4 to generate the first \textit{Thought} step and select the correct action for the general questions.

After supervised fine-tuning, Llama2-7B's performance improves by an average of 14\%, and incorrect tool use drops by 16.8\% in general datasets. Specifically, in the PIQA, accuracy in the \textit{Search Engine} improves by 40\%, and the \textit{Calculator} call rate decreases by 74\%, as detailed in Table~\ref{tab:finetuning}.


\section{Conclusion}

In this paper, we explore whether LLMs can discern their ability boundaries and use tools flexibly. We introduce the WTU-Eval to assess LLMs with eleven datasets and four tools. The results of WTU-Eval reveal that LLMs frequently struggle to determine tool use in general datasets, and their performance in tool-usage datasets improves when their ability is similar to ChatGPT. In both datasets, incorrect tool usage significantly impairs LLMs' performance. After detailed analysis, we also introduce a dataset focused on improving decision-making in tool usage, which successfully enhances Llama2-7B's performance and reduces unnecessary tool invocations.

Our work points out the overlooked shortcomings in tool usage by LLMs, i.e., they struggle to recognize their capability boundaries and lack of tool usage decision-making capability. We use the WTU-Eval to test eight LLMs, which is the first benchmark to evaluate whether LLMs can use tools accurately. Future works include adding more datasets and tools, and testing more types of LLMs.

\section*{Limitations}
This study's limitations arise from computational constraints, limiting our model selection to exclude larger variants like Llama2-70B, and from the models' slow processing of tool directives, leading us to evaluate a sampled subset of the test set, potentially causing result discrepancies with other studies.

\section*{Ethics Statement and Broader Impacts}

This study exclusively utilized datasets and toolsets that are publicly available and previously published, ensuring they contain no offensive or harmful content. We rigorously adhere to ethical standards, including a thorough review of materials to safeguard privacy and integrity.  

This study is pivotal for the practical application of LLMs, as it aims at reducing unnecessary tool invocations, thereby enhancing the efficiency of tool usage. This optimization in tool interaction not only advances the development of AI but also ensures more effective and streamlined AI operations, leading to smarter and more efficient AI systems that better serve the needs across different sectors and research disciplines.

\bibliography{anthology,custom}
\bibliographystyle{acl_natbib}

\appendix
\section{Hardware Configuration and Datasets}\label{dataset}

In this study, the hardware configuration comprises an NVIDIA GeForce RTX 3090 GPU with 20 GB of memory for a tool-usage task that costs several minutes, and an NVIDIA A100 GPU with 80 GB of memory for the fine-tuning task, requiring six hours to complete nine epochs tuning.

Table~\ref{tab:datasets} outlines each dataset's size, action type, and tool call steps. Test sizes are chosen based on reasoning complexity and time costs, leading to random sampling from original test sets, which may cause different results from some public benchmarks. Action types relate to tool interactions, and call steps indicate if a task requires a single or multiple tool call step. 

\begin{table}[t]
\small
\centering
\resizebox{1.0\linewidth}{!}{
\begin{tabular}{lrccccc}
\toprule
\textbf{Test Set} & \textbf{Test Size} & \textbf{Action Type} & \textbf{Call Step} \\
\midrule
\multicolumn{2}{l}{\textbf{Tool Dataset}} & & & & \\
\midrule
MLQA(\citeyear{lewis2019mlqa}) & 96 & Single & Single \\
ASDiv(\citeyear{miao2021diverse}) & 300 & Single & Multiple \\
GSM8K(\citeyear{cobbe2021training}) & 100 & Single & Multiple \\
MathQA(\citeyear{amini2019mathqa}) & 100 & Single & Multiple \\
HotpotQA(\citeyear{yang2018hotpotqa}) & 200 & Multiple & Multiple \\
RealtimeQA(\citeyear{kasai2022realtime}) & 120 & Multiple & Multiple \\
\midrule
\multicolumn{2}{l}{\textbf{General Dataset}} & & & & & \\
\midrule
BoolQ(\citeyear{clark2019boolq}) & 100 & - & - \\
RACE(\citeyear{lai2017race}) & 100 & - & - \\
PIQA(\citeyear{bisk2020piqa}) & 100 & - & - \\
RTE\footnote{\url{https://aclweb.org/aclwiki/Recognizing_Textual_Entailment}} & 100 & - & - \\
HellaSwag(\citeyear{zellers2019HellaSwag}) & 100 & - & - \\
\bottomrule
\end{tabular}
}
\caption{Details of all datasets evaluated in WTU-Eval, include the type of tool action, and tool call steps involved. Action Type and Tool Call Step are not listed for general datasets as they theoretically involve only a single step.}
\label{tab:datasets}
\end{table}

\section{Examples of Successful Types}

\begin{itemize}
    \item \textbf{Correct Tool Usage Example of Calculator}\\
    Task Description:\\
    Given a context and related question, try your best to answer the question.
    · Context: Being his favorite, he saved checking on the grapevines for his last stop. He was told by one the pickers that they fill 324 drums of grapes per day.\\
    ·Question: How many drums of grapes would be filled in 9 days?\\
    \textit{Thought}: I should use calculator for Multiplication to finish this task.\\
    \textit{Action}: Calculator\\
    \textit{Action Input}: 324*9\\
    \textit{Observation}: \{`@title': `Input', `subpod': \{`@title': `', `img': \{`@src':\{given link\}, `@alt': `324 * 9', `@title': `324 * 9', `@contenttype': `image/gif'\}, `plaintext': `324 * 9'\}\}, \{`@title': `Result', `@primary': `true', `subpod': \{`@title': `', `img': \{`@src':\{given link\}, `@alt': `2916', `@title': `2916', `@contenttype': `image/gif'\}, `plaintext': `2916'\}, `states': \{`@count': `1', `state': \{`@name': `Step-by-step solution', `@input': `Result\_Step-by-step solution', `@stepbystep': `true', `@buttonstyle': `StepByStepSolution'\}\}\}\\
    \textit{Thought:} I think I got the final answer.\\
    \textit{Action:} Finish\\
    \textit{Action Input:} 2916 (drums)\\
    \textit{Final Answer:} 2916 (drums)"\\
    
    \item \textbf{Correct Tool Usage Example of Machine Translator}\\
    Task Description:\\
    Given an English context and related Arabic question (you can translate the question into English), answer the question in English.\\
    English Context: In 1884, a paper was signed at Togoville with the King Mlapa III, whereby Germany claimed a protectorate over a stretch of territory along the coast and gradually extended its control inland. Its borders were defined after the capture of hinterland by German forces and signing agreements with France and Britain. In 1905, this became the German colony of Togoland. The local population was forced to work, cultivate cotton, coffee and cocoa and pay high taxes. A railway and the port of Lom\u00e9 were built for export of agricultural products. The Germans introduced modern techniques of cultivation of cocoa, coffee and cotton and developed the infrastructure.\\
    · Arabic Question:\begin{RLtext}
        متى تم توقيع الأوراق؟
    \end{RLtext}\\
    \textit{Thought:} I need to translate the question \begin{RLtext}
        متى تم توقيع الأوراق؟
    \end{RLtext}\\

    \textit{Action:} Translator\\
    \textit{Action Input:}\begin{RLtext}
        متى تم توقيع الأوراق؟
    \end{RLtext}\\
    \textit{Observation:} When were these documents signed?\\
    \textit{Thought:} I think I got the final answer\\
    \textit{Action:} Finish\\
    \textit{Action Input:} 1884\\
    \textit{Final Answer:} 1884\\

    \item \textbf{Correct Tool Usage Example of Search Engine}\\
    Task Description: Choose the best option to answer the given question\\
    · Question: Which country declared a National State of Emergency this week for the third time in its history due to a major storm?\\
    · Option:['Greece', 'New Zealand', 'Mexico', 'Finland']\\
    \textit{Thought:} To answer the question, I can use the BingSearch API to search for news articles specifically about the given options: Greece, New Zealand, Mexico, and Finland, along with keywords like \"national state of emergency\" and \"major storm\".\\
    \textit{Action:} BingSearch\\
    \textit{Action Input:} 'Greece national state of emergency major storm'\\
    \textit{Observation:} page: 1\\
    title: Storm Elias: Greece hit by several months of rain in a day, just ... - CNN\\summary: CNN Storm Elias dumped several months-worth of rain in less than a day as it swept across Greece on Wednesday, just weeks after Storm Daniel killed 17 people in the southern European nation.\\page: 2\\title: Greek Government Holds Emergency Meeting As Dangerous Storm Front Set ...\\summary: An emergency weather forecast issued on Friday by the Hellenic National Meteorological Service (EMY) speaks of heavy rainfall and storms over western, central and northern Greece starting Monday, before spreading to more areas by mid-next week. Tags: greece, Greek, Greek news, greek storm, Vassilis Kikilias Previous Article\\page: 3\\title: Greek floods and fires expose Europe's frail climate defences\\summary: Following the previous major storm that flooded Thessaly in 2020, Greece's conservative government promised to prevent a repeat of the disaster. ... Greece's emergency plan issued in October 2022 ...\\
    \textit{Thought:} Based on the search results, Greece has been hit by several major storms, with heavy rainfall and floods causing significant damage. The articles mention a national state of emergency being declared due to these storms, which aligns with the question. Therefore, the answer to the question is Greece.
    \textit{Action:} Finish\\
    \textit{Action Input:} Greece\\
    \textit{Final Answer:} Greece\\
    
    \item \textbf{Correct Tool Usage Example of Wikipedia Search}\\
    Task Description:\\
    Answer the given question in a short sentence:\\
    Were Michael Winner and Guy Hamilton both English film directors?\\
    \textit{Thought:} I need to find information about Michael Winner and Guy Hamilton to determine if they were both English film directors.\\
    \textit{Action:} WikiSearch\\
    \textit{Action Input:} \"Michael Winner\"\
    \textit{Observation:} Robert Michael Winner (30 October 1935 \u2013 21 January 2013) was a British filmmaker, writer, and media personality. He is known for directing numerous action, thriller, and black comedy films in the 1960s, 1970s and 1980s, including several collaborations with actors Oliver Reed and Charles Bronson.[1][2]. Winner's best-known works include Death Wish (1974) and its first two sequels, the World War II comedy Hannibal Brooks (1969), the hitman thriller The Mechanic (1972), the supernatural horror film The Sentinel (1977), the neo-noir The Big Sleep (1978), the satirical comedy Won Ton Ton, the Dog Who Saved Hollywood (1976), and the Revisionist Westerns Lawman (1971) and Chato's Land (1972).. Winner was known as a media personality in the United Kingdom, appearing regularly on television talk programmes and publishing a restaurant review column for The Sunday Times. He was also a founder of the Police Memorial Trust..\\
    \textit{Thought:} I found information about Michael Winner, but I still need to find information about Guy Hamilton.\\
    \textit{Action:} WikiSearch\\
    \textit{Action Input:} \"Guy Hamilton\"\\
    \textit{Observation:} Mervyn Ian Guy Hamilton, DSC (16 September 1922 \u2013 20 April 2016) was an English film director. He directed 22 films from the 1950s to the 1980s, including four James Bond films.. Hamilton was born in Paris on 16 September 1922, son of Frederick William Guy Hamilton (1895-1988), press attach\u00e9 to the British embassy in Paris and Captain in the King's Own Royal Lancaster Regiment, and Winifred Grace Culling (1895-1970), daughter of William Archibald Culling Fremantle, of the Church Missionary Society in India.[1] His mother was a great-granddaughter of the Christian campaigner Sir Culling Eardley, 3rd Baronet, and of the politician Thomas Fremantle, 1st Baron Cottesloe. His parents divorced in 1923, and Hamilton attended school in England.[2]. [3] His first exposure to the film industry came in 1938, when he was a clapperboard boy at the Victorine Studios in Nice.\\
    \textit{Thought:} I have found information about both Michael Winner and Guy Hamilton. Now I need to determine if they were both English film directors.\\
    \textit{Action:} Finish\\
    \textit{Action Input:} Yes, they were both English film directors.\\
    \textit{Final Answer:} Yes, they were both English film directors.

    \item \textbf{Correct Example of General Datasets}\\
    Task Description:\\
Given a passage and a question, try your best to answer the question true or false according to the passage.\\
· Passage: In statistics, ordinary least squares (OLS) is a type of linear least squares method for estimating the unknown parameters in a linear regression model. OLS chooses the parameters of a linear function of a set of explanatory variables by the principle of least squares: minimizing the sum of the squares of the differences between the observed dependent variable (values of the variable being predicted) in the given dataset and those predicted by the linear function.
· Question: are least squares the same as linear regression?
\textit{Thought:} I can answer the question without any tools. According to the passage, least squares are not the same as linear regression.  Rather, least squares is a method used within linear regression for estimating the unknown parameters.  \\
\textit{Action:} Finish\\
\textit{Action Input:} false\\
\textit{Final Answer:} false\\
\end{itemize}

\section{Examples of Failure Type}
\begin{itemize}
    \item \textbf{Incorrect or Unnecessary Tool Invocation}\\ \textbf{Model: ChatGPT}\\
    · Context: \\English breakfast is a very big meal--eggs, tomatoes, tea, coffee... For many people, lunch is a quick meal. In cities, there are a lot of sandwich bars. People can buy sandwiches there. Students can have a hot meal at school, but many just take a sandwich, a drink, and some fruit from home. "Tea" means two things. It is a drink and a meal! Some people have afternoon tea, with sandwiches, cakes, and a cup of tea. They usually have dinner quite early, between 6:00 and 8:00(......), and often all the family eat together.\\
    People often get take-away meals--they buy the food outside\\
    · Question: \\When they get a takeaway meal, they often eat it ()?\\
    · Options:\\A. "at home"\\B. "in the school"\\C. "outside"\\D. "in the bars"\\
    \textit{Thought:} There is no answer in the passage.\\
    \textit{Action:} BingSearch\\
    \textit{Action Input:} Where do people often eat takeaway meals?\\
    \textit{Error Raise:} It's general data, you can not use the tools.\\
    \item \textbf{Null Content}\\ \textbf{Model: Llama2-13B-hf}\\\textbf{Thought Null}\\· Context:\\
  Olivia picked up sixteen pieces of paper from the floor. If Edward picked up three pieces,\\
· Question:  how many did they pick up in total?\\
\textit{Thought:} I should try to calculate the total number of pieces of paper picked up.\\
\textit{Action:} Calculator\\
\textit{Action Input:} 16 + 3\\
\textit{Observation:} \{'@title': 'Input', 'subpod': \{'@title': '', 'img': \{'@src':\{given link\}, '@alt': '16 + 3', '@title': '16 + 3', '@contenttype': 'image/gif'\}, 'plaintext': '16 + 3'\}\}, \{'@title': 'Result', '@primary': 'true', 'subpod': \{'@title': '', 'img': \{'@src':\{given link\}, '@alt': '19', '@title': '19', '@contenttype': 'image/gif'\}, 'plaintext': '19'\}, 'states': \{'@count': '1', 'state': \{'@name': 'Step-by-step solution', '@input': 'Result\_Step-by-step solution', '@stepbystep': 'true', '@buttonstyle': 'StepByStepSolution'\}\}\}\\
\textit{Thought:}\\
\textit{Error Raise:} Thought string is null, model refused to answer\\
\textbf{Action Null}\\
· Context: I have 3 hundreds, 8 tens, and 3 ones.\\
· Question: What number am I?\\
\textit{Thought:} I should try to find a way to convert the numbers into a number that I can add.\\
\textit{Action:} \\
\textit{Error Raise:} Action string is null, model refused to answer\\
\textbf{Action Input Null} \\
· Context:\\
  Three carpenters are installing some doors for a building.\\
  Carpenter A installs a door every 40 min.\\
  Carpenter B installs a door every 1 h 20 min.\\
  Carpenter C installs a door every 1 h 40 min.\\
  If they start installing three doors at the same time at 9:15,\\
· Question:\\
  at what time will they install doors at the same again?\\
\textit{Thought:} I can use the calculator to calculate the time.\\
\textit{Action:} Calculator\\
\textit{Error Raise:}
Action Input string is null, model refused to answer\\
    \item \textbf{Recall correct tool not reasoning} \\ \textbf{Model: Llama2-13B-hf}\\
    · Context:\\ 
  Jerry was weighing boxes for moving. The first was 2 pounds, the second was 11 pounds and the last was 5 pounds.\\
· Question:\\
  What is the combined weight of all three boxes?\\ 
\textit{Thought:} I need to calculate the combined weight of the boxes.\\
\textit{Action:} Calculator\\
\textit{Action Input:} 2 + 11 + 5\\
\textit{Observation:}  \{'@title': 'Result', '@primary': 'true', 'subpod': \{'@title': '', 'img': \{'@src':\{given link\}, '@alt': '18', '@title': '18', '@contenttype': 'image/gif'\}\\
\textit{Thought:} I need to calculate the combined weight of the boxes.\\
\textit{Action:} Calculator\\
\textit{Action Input:} 2 + 11 + 5\\
\textit{Observation:} \{'@title': 'Result', '@primary': 'true', 'subpod': \{'@title': '', 'img': \{'@src':\{given link\}, '@alt': '18', '@title': '18', '@contenttype': 'image/gif'\}\\
\textit{Thought:} I need to calculate the combined weight of the boxes.\\
\textit{Action:} Calculator\\
\textit{Action Input:} 2 + 11 + 5\\
\textit{Observation:} \{'@title': 'Result', '@primary': 'true', 'subpod': \{'@title': '', 'img': \{'@src':\{given link\}, '@alt': '18', '@title': '18', '@contenttype': 'image/gif'\}\\
\textit{Thought:} I need to calculate the combined weight of the boxes.\\
\textit{Action:} Calculator\\
\textit{Action Input:} 2 + 11 + 5\\
\textit{Error Raise:} Continue to try the same tool and parameter, model refused to answer\\
    \item \textbf{Recall invalid tool after reminding} \\ \textbf{Model: Zephyr-7B} \\
    · English Context:\\
  "Das aschkenasische Rabbinat : Studien über Glaube und Schicksal" (Julius Carlebach) / Die Rolle des Rabbiners in Deutschland heute (Ahron Daum)\\
· Arabic Question:\\
 \begin{RLtext}ماذا كان الاسم الأخير لأهرون؟\end{RLtext}\\
\textit{Thought:} I can translate the question into English and then use the translator to answer the question.\\
\textit{Action:} Translate the question into English.\\
\textit{Action Input:} \begin{RLtext}ماذا كان الاسم الأخير لأهرون؟\end{RLtext}\\
\textit{Observation:} Please choose a valid tool in Translator!\\
\textit{Thought:} I can translate the question into English and then use the translator to answer the question.\\
\textit{Action:} Translate the question into English\\
\textit{Action Input:}\begin{RLtext}ماذا كان الاسم الأخير لأهرون؟\end{RLtext}\\
\textit{Observation:} Please choose a valid tool in Translator!\\
\textit{Thought:} I can translate the question into English and then use the translator to answer the question.\\
\textit{Action:} Translate the question into English\\
\textit{Action Input:}\begin{RLtext}ماذا كان الاسم الأخير لأهرون؟\end{RLtext}\\
\textit{Observation:} Please choose a valid tool in Translator!\\
\textit{Error Raise:} Continue to use the invalid action tool, the model refused to answer.\\
    \item \textbf{Similar invalid parameter recycle} \\ \textbf{Model: Llama2-13B-Chat-hf}\\
    · Context:\\
    Uncle Dave bought 143 ice cream sandwiches. If he wants to give them to his 11 hungry nieces,\\
    · Question:\\
    how many can each niece get?\\
    \textit{Thought:} Hmm, I wonder if I can use the calculator to calculate the number of ice cream sandwiches each niece can get.\\
    \textit{Action:} Calculator\\
    \textit{Action Input:} 143 ice cream sandwiches / 11 nieces\\
    \textit{Observation:} WolframAlpha API cannot parse the input query.\\
    \textit{Thought:} Oh no! Maybe I need to rephrase the question.\\
    \textit{Action:} Calculator\\
    \textit{Action Input:} 143 ice cream sandwiches / 11 nieces = how many ice cream sandwiches per niece?\\
    \textit{Observation:} WolframAlpha API cannot parse the input query.\\
    \textit{Thought:} Hmm, maybe I need to convert the number of ice cream sandwiches to a more manageable unit.\\
    \textit{Action:} Calculator\\
    \textit{Action Input:} 143 ice cream sandwiches / 11 nieces = how many ice cream sandwiches per niece (in dozens)?\\
    \textit{Observation:} WolframAlpha API cannot parse the input query.\\
    \textit{Error Raise:} Observation return same content, model refused to answer.
    \item \textbf{Endless retry}\\ \textbf{Model: Zephyr-7B}\\
    Question: The McCarran Airport Connector serves what airport named after late U.S. Senator Pat McCarran?\\
    \textit{Thought:} I need to find information about the McCarran Airport Connector and the airport it serves, which is named after a late U.S. Senator.\\
    \textit{Action:} WikiSearch\\
    \textit{Action Input:} McCarran Airport Connector\\
    \textit{Observation:} State Route 171. The Harry Reid Airport Connector (RAC) is a limited-access roadway system located in Paradise, an unincorporated town in the Las Vegas Valley, Clark County, Nevada, United States. Composed of State Route 171 (SR 171), the Airport Tunnel and arterial streets, the airport connector provides vehicular access to the passenger terminals at Harry Reid International Airport. Despite being completely owned by Clark County, the first 0.685 miles (1.102 km) of the Harry Reid Airport Connector is maintained by NDOT as unsigned SR 171, while the remaining section is maintained by Clark County.[1]. The Harry Reid Airport Connector was constructed and opened to traffic in 1994, in conjunction with the completion of I-215 between Interstate 15 and Warm Springs Road.\\
    \textit{Thought:} I have found information about the McCarran Airport Connector, but I need to find information about the airport it serves, which is named after a late U.S. Senator.\\
    \textit{Action:} WikiSearch\\
    \textit{Action Input:} Pat McCarran\\
    \textit{Observation:} Patrick Anthony McCarran (August 8, 1876 – September 28, 1954) was an American farmer, attorney, judge, and Democratic politician who represented Nevada in the United States Senate from 1933 until 1954. McCarran was born in Reno, Nevada, attended Nevada State University (now the University of Nevada, Reno), and was a farmer and rancher. In 1902, he won election to the Nevada Assembly but left office in 1905 after an unsuccessful campaign for the Nevada State Senate. He studied law privately and was admitted to the bar in 1905, then won election as Nye County District Attorney. He served a two-year term, after which he returned to Reno.\\
    \textit{Error Raise:} The text is too long
\label{appendix:ErrorType}
\end{itemize}

\section{Prompts}
\label{prompts}
The prompts used in the evaluation are shown in this section.

\begin{figure}[H]
  \centering
   \resizebox{\linewidth}{!}{\includegraphics{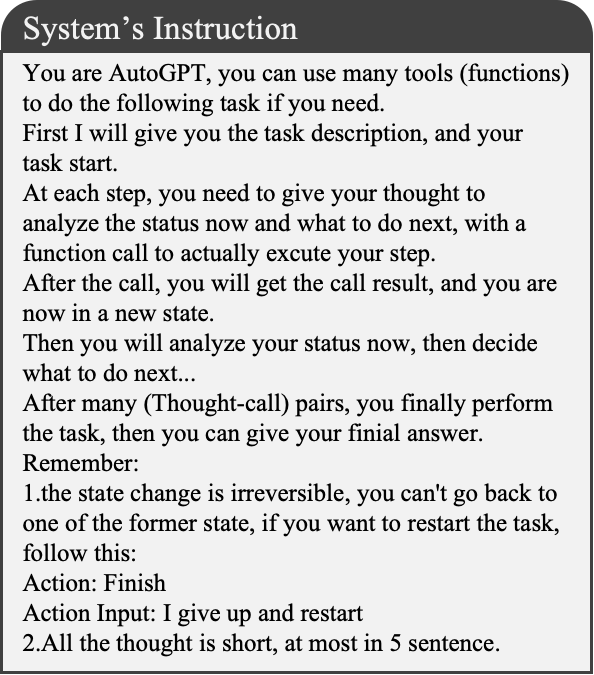}}\vspace{5pt}\\
    \resizebox{\linewidth}{!}
{\includegraphics{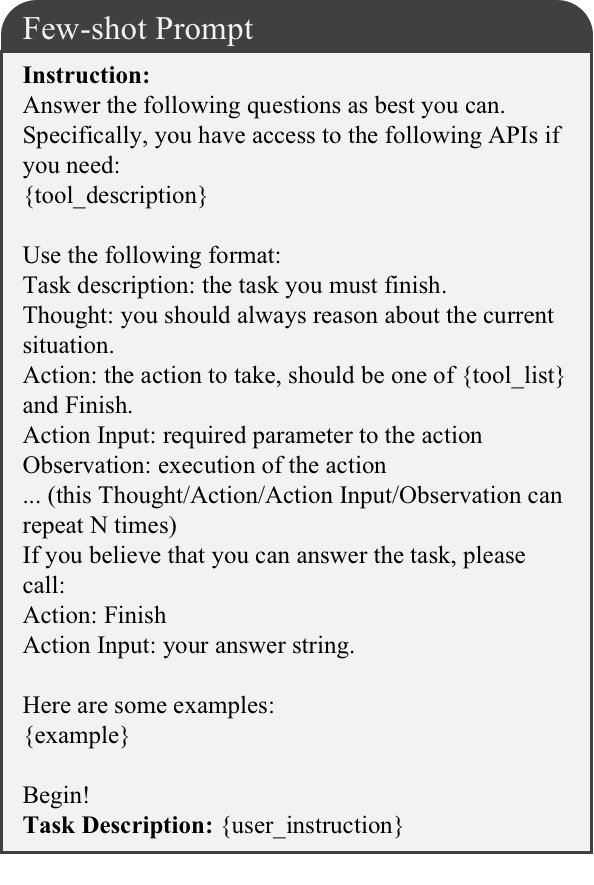}}\vspace{5pt}\\

\end{figure}

\begin{figure}[H]
  \centering
\resizebox{\linewidth}{!}{\includegraphics{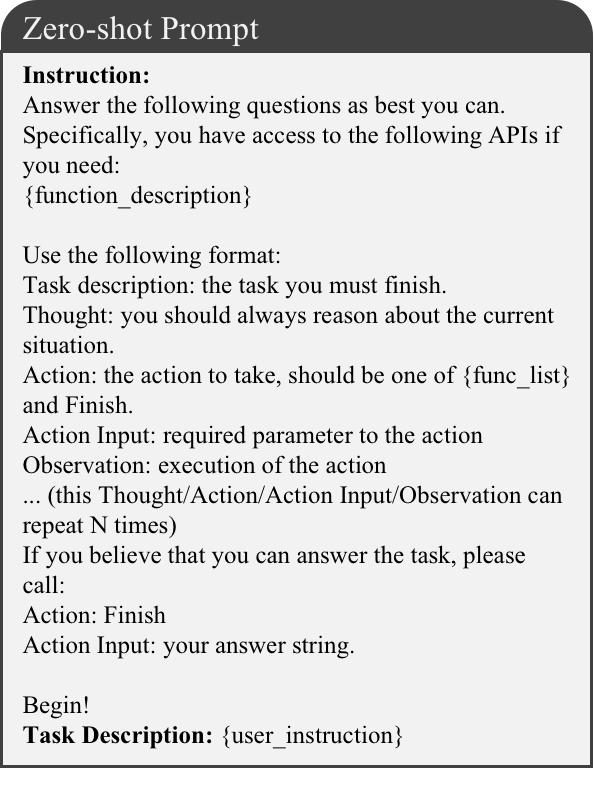}}\vspace{5pt}\\

     \resizebox{\linewidth}{!}{\includegraphics{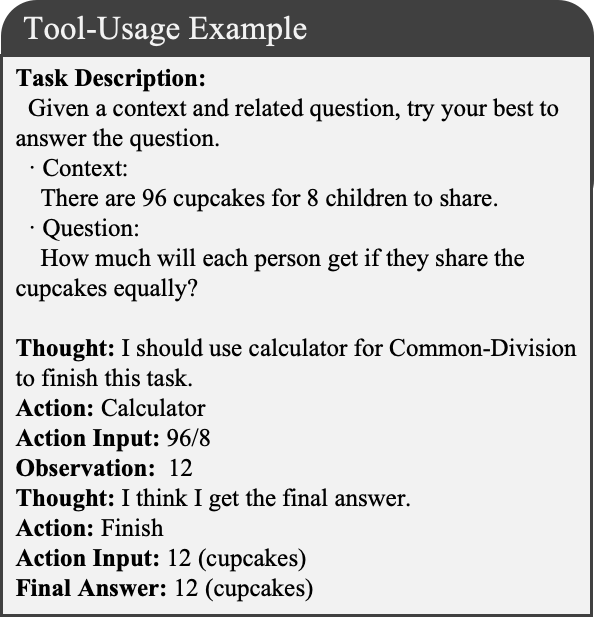}}\vspace{5pt}\\

\end{figure}

\begin{figure}[H]
  \centering 
    \resizebox{\linewidth}{!}{\includegraphics{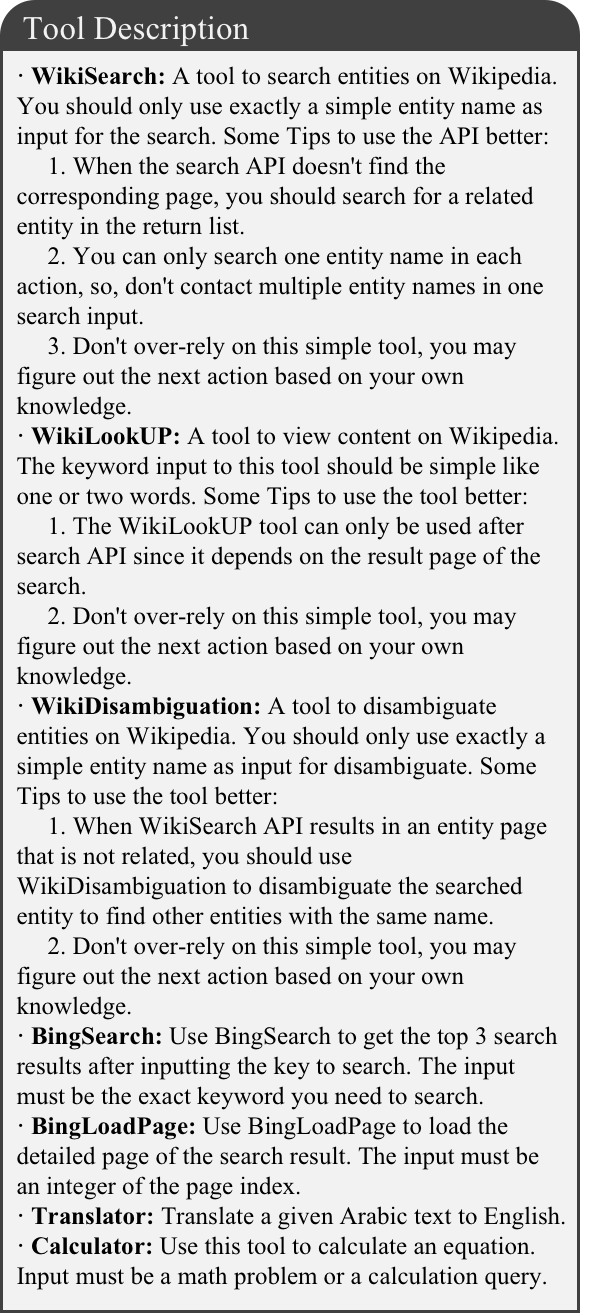}}\vspace{5pt}\\
    \resizebox{\linewidth}{!}{\includegraphics{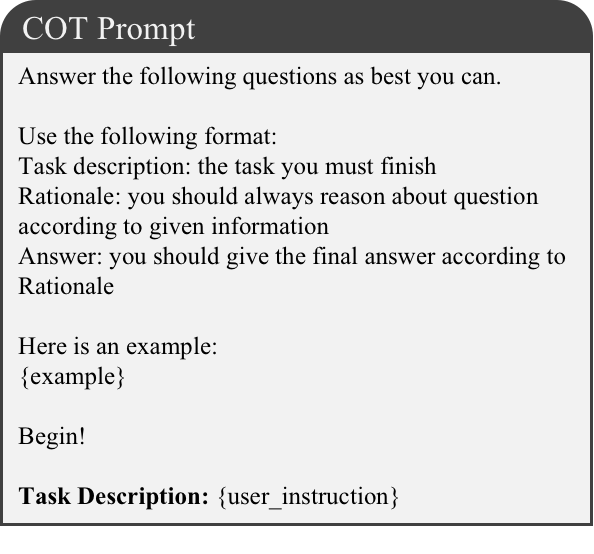}}\vspace{5pt}\\
\end{figure}

\begin{figure}[H]
  \centering
  
    \resizebox{\linewidth}{!}{\includegraphics{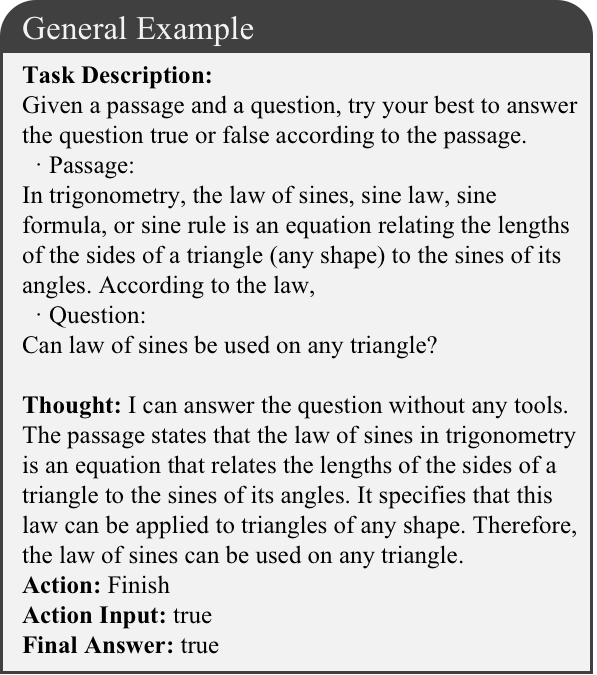}}\vspace{5pt}\\
    \resizebox{\linewidth}{!}{\includegraphics{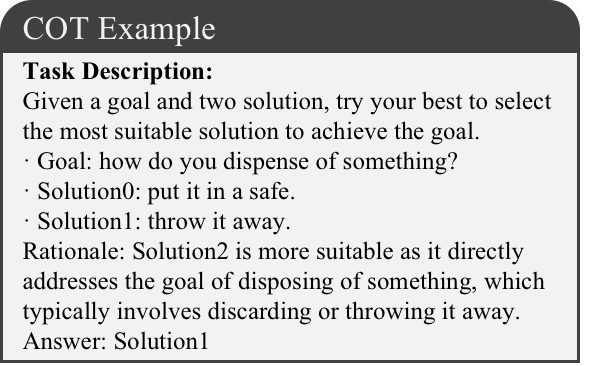}}\vspace{5pt}\\
  
\end{figure}

\onecolumn
\section{Tool Impact on Model Performance across General Datasets}\label{all_fs_zs_general}

\begin{table}[H]
\small 
\resizebox{\linewidth}{!}{
\begin{tabular}{llrrrrrrrr}

\label{tab:few-shot_and_zero-shot} \\
\toprule
\multirow{3}{*}{\textbf{Test set}} & \multirow{3}{*}{\textbf{Tool}} & \multicolumn{8}{c}{\textbf{Model w/ Tool}} \\
 & & \textbf{GPT3} & \textbf{ChatGPT} & \multicolumn{2}{c}{\textbf{Llama2-7B}} & \multicolumn{2}{c}{\textbf{Llama2-13b}} & \textbf{Zephyr-7B} & \textbf{Chatglm3-6b} \\
 & & & & \textbf{Base} & \textbf{Chat} & \textbf{Base} & \textbf{Chat} & & \\

\midrule
\multirow{12}{*}{Boolq} & \multirow{2}{*}{Translator} & \cellcolor{yellow!50}68.00   & \cellcolor{yellow!50}80.00   & \cellcolor{yellow!50}64.00   & \cellcolor{yellow!50}45.00   & \cellcolor{yellow!50}62.00   & \cellcolor{yellow!50}77.00   & \cellcolor{yellow!50}52.00   & \cellcolor{yellow!50}31.00  \\
 & & 56.00  & 11.00  & 0.00  & 0.00  & 0.00  & 0.00  & 35.00  & 10.00 \\
 & \multirow{2}{*}{Calculator} & \cellcolor{yellow!50}83.00   & \cellcolor{yellow!50}85.00   & \cellcolor{yellow!50}59.00   & \cellcolor{yellow!50}42.00   & \cellcolor{yellow!50}56.00   & \cellcolor{yellow!50}52.00   & \cellcolor{yellow!50}8.00   & \cellcolor{yellow!50}43.00  \\
 & & 47.00  & 5.00  & 0.00  & 0.00  & 0.00  & 0.00  & 33.00  & 7.00 \\
 & \multirow{2}{*}{Search Engine} & \cellcolor{yellow!50}45.00   & \cellcolor{yellow!50}70.00   & \cellcolor{yellow!50}41.00   & \cellcolor{yellow!50}9.00   & \cellcolor{yellow!50}46.00   & \cellcolor{yellow!50}60.00   & \cellcolor{yellow!50}53.00   & \cellcolor{yellow!50}23.00  \\
 & & 31.00  & 8.00  & 0.00  & 0.00  & 0.00  & 1.00  & 35.00  & 8.00 \\
 & \multirow{2}{*}{WikiPedia Search}& \cellcolor{yellow!50}38.00   & \cellcolor{yellow!50}70.00   & \cellcolor{yellow!50}56.00   & \cellcolor{yellow!50}34.00   & \cellcolor{yellow!50}52.00   & \cellcolor{yellow!50}56.00   & \cellcolor{yellow!50}77.00   & \cellcolor{yellow!50}14.00  \\
 & & 42.00  & 8.00  & 0.00  & 0.00  & 0.00  & 0.00  & 34.00  & 18.00 \\
 & \multirow{2}{*}{All} & \cellcolor{yellow!50}58.50   & \cellcolor{yellow!50}76.25   & \cellcolor{yellow!50}55.00   & \cellcolor{yellow!50}32.50   & \cellcolor{yellow!50}54.00   & \cellcolor{yellow!50}61.25   & \cellcolor{yellow!50}47.50   & \cellcolor{yellow!50}27.75  \\
 & & 20.00  & 6.00  & 0.00  & 2.00  & 0.00  & 0.00  & 17.00  & 20.00 \\
 \midrule 
 \multirow{12}{*}{RACE} & \multirow{2}{*}{Translator} & \cellcolor{yellow!50}86.93   & \cellcolor{yellow!50}79.47   & \cellcolor{yellow!50}71.47   & \cellcolor{yellow!50}40.00   & \cellcolor{yellow!50}70.07   & \cellcolor{yellow!50}58.67   & \cellcolor{yellow!50}55.20   & \cellcolor{yellow!50}59.73  \\
 & & 52.00  & 27.00  & 5.00  & 0.00  & 0.00  & 0.00  & 29.00  & 22.00 \\
 & \multirow{2}{*}{Calculator} & \cellcolor{yellow!50}86.93   & \cellcolor{yellow!50}76.26   & \cellcolor{yellow!50}69.60   & \cellcolor{yellow!50}67.20   & \cellcolor{yellow!50}40.26   & \cellcolor{yellow!50}35.20   & \cellcolor{yellow!50}40.80   & \cellcolor{yellow!50}74.13  \\
 & & 57.00  & 28.00  & 6.00  & 0.00  & 0.00  & 1.00  & 33.00  & 30.00 \\
 & \multirow{2}{*}{Search Engine} & \cellcolor{yellow!50}78.13   & \cellcolor{yellow!50}77.60   & \cellcolor{yellow!50}41.60   & \cellcolor{yellow!50}40.26   & \cellcolor{yellow!50}78.40   & \cellcolor{yellow!50}62.13   & \cellcolor{yellow!50}74.13   & \cellcolor{yellow!50}62.93  \\
 & & 25.00  & 23.00  & 0.00  & 0.00  & 0.00  & 1.00  & 37.00  & 32.00 \\
 & \multirow{2}{*}{WikiPedia Search}& \cellcolor{yellow!50}79.73   & \cellcolor{yellow!50}76.53   & \cellcolor{yellow!50}50.93   & \cellcolor{yellow!50}54.13   & \cellcolor{yellow!50}59.47   & \cellcolor{yellow!50}54.13   & \cellcolor{yellow!50}67.47   & \cellcolor{yellow!50}67.47  \\
 & & 30.00  & 33.00  & 0.00  & 0.00  & 0.00  & 0.00  & 51.00  & 21.00 \\
 & \multirow{2}{*}{All} & \cellcolor{yellow!50}82.93   & \cellcolor{yellow!50}77.47   & \cellcolor{yellow!50}58.40   & \cellcolor{yellow!50}50.40   & \cellcolor{yellow!50}62.05   & \cellcolor{yellow!50}52.53   & \cellcolor{yellow!50}59.40   & \cellcolor{yellow!50}66.06  \\
 & & 6.00  & 30.00  & 14.00  & 0.00  & 0.00  & 0.00  & 17.00  & 22.00 \\
 \midrule 
  \multirow{12}{*}{PIQA} & \multirow{2}{*}{Translator} & \cellcolor{yellow!50}77.00   & \cellcolor{yellow!50}57.00   & \cellcolor{yellow!50}51.00   & \cellcolor{yellow!50}41.00   & \cellcolor{yellow!50}41.00   & \cellcolor{yellow!50}48.00   & \cellcolor{yellow!50}38.00   & \cellcolor{yellow!50}37.00  \\
 & & 63.00  & 40.00  & 5.00  & 0.00  & 0.00  & 0.00  & 23.00  & 1.00 \\
 & \multirow{2}{*}{Calculator} & \cellcolor{yellow!50}68.00   & \cellcolor{yellow!50}61.00   & \cellcolor{yellow!50}14.00   & \cellcolor{yellow!50}47.00   & \cellcolor{yellow!50}42.00   & \cellcolor{yellow!50}32.00   & \cellcolor{yellow!50}3.0   & \cellcolor{yellow!50}21.00  \\
 & & 51.00  & 40.00  & 0.00  & 0.00  & 0.00  & 0.00  & 18.00  & 10.00 \\
 & \multirow{2}{*}{Search Engine} & \cellcolor{yellow!50}25.00   & \cellcolor{yellow!50}55.00   & \cellcolor{yellow!50}0.00   & \cellcolor{yellow!50}19.00   & \cellcolor{yellow!50}25.00   & \cellcolor{yellow!50}44.00   & \cellcolor{yellow!50}3.00   & \cellcolor{yellow!50}18.00  \\
 & & 13.00  & 44.00  & 0.00  & 0.00  & 0.00  & 0.00  & 49.00  & 5.00 \\
 & \multirow{2}{*}{WikiPedia Search}& \cellcolor{yellow!50}31.00   & \cellcolor{yellow!50}62.00   & \cellcolor{yellow!50}10.00   & \cellcolor{yellow!50}17.00   & \cellcolor{yellow!50}1.00   & \cellcolor{yellow!50}50.00   & \cellcolor{yellow!50}19.00   & \cellcolor{yellow!50}0.00  \\
 & & 26.00  & 40.00  & 0.00  & 0.00  & 0.00  & 0.00  & 47.00  & 6.00 \\
 & \multirow{2}{*}{All} & \cellcolor{yellow!50}50.25   & \cellcolor{yellow!50}58.75   & \cellcolor{yellow!50}18.75   & \cellcolor{yellow!50}31.00   & \cellcolor{yellow!50}27.25   & \cellcolor{yellow!50}43.50   & \cellcolor{yellow!50}15.75   & \cellcolor{yellow!50}19.00  \\
 & & 3.00  & 39.00  & 14.00  & 0.00  & 0.00  & 0.00  & 4.00  & 3.00 \\
 \midrule 
  \multirow{12}{*}{RTE} & \multirow{2}{*}{Translator} & \cellcolor{yellow!50}75.00   & \cellcolor{yellow!50}51.00   & \cellcolor{yellow!50}51.00   & \cellcolor{yellow!50}44.00   & \cellcolor{yellow!50}34.00   & \cellcolor{yellow!50}36.00   & \cellcolor{yellow!50}15.00   & \cellcolor{yellow!50}46.00  \\
 & & 31.00  & 21.00  & 0.00  & 0.00  & 0.00  & 0.00  & 45.00  & 21.00 \\
 & \multirow{2}{*}{Calculator} & \cellcolor{yellow!50}71.00   & \cellcolor{yellow!50}60.00   & \cellcolor{yellow!50}53.00   & \cellcolor{yellow!50}32.00   & \cellcolor{yellow!50}7.00   & \cellcolor{yellow!50}42.00   & \cellcolor{yellow!50}3.0   & \cellcolor{yellow!50}47.00  \\
 & & 36.00  & 29.00  & 2.00  & 0.00  & 0.00  & 0.00  & 41.00  & 26.00 \\
 & \multirow{2}{*}{Search Engine} & \cellcolor{yellow!50}61.00   & \cellcolor{yellow!50}45.00   & \cellcolor{yellow!50}48.00   & \cellcolor{yellow!50}13.00   & \cellcolor{yellow!50}13.00   & \cellcolor{yellow!50}34.00   & \cellcolor{yellow!50}2.00   & \cellcolor{yellow!50}42.00  \\
 & & 10.00  & 22.00  & 0.00  & 1.00  & 0.00  & 0.00  & 28.00  & 37.00 \\
 & \multirow{2}{*}{WikiPedia Search}& \cellcolor{yellow!50}58.00   & \cellcolor{yellow!50}44.00   & \cellcolor{yellow!50}30.00   & \cellcolor{yellow!50}4.00   & \cellcolor{yellow!50}0.00   & \cellcolor{yellow!50}26.00   & \cellcolor{yellow!50}42.00   & \cellcolor{yellow!50}50.00  \\
 & & 10.00  & 15.00  & 0.00  & 0.00  & 0.00  & 0.00  & 28.00  & 35.00 \\
 & \multirow{2}{*}{All} & \cellcolor{yellow!50}66.25   & \cellcolor{yellow!50}50.00   & \cellcolor{yellow!50}45.50   & \cellcolor{yellow!50}23.25   & \cellcolor{yellow!50}13.50   & \cellcolor{yellow!50}34.50   & \cellcolor{yellow!50}15.50   & \cellcolor{yellow!50}46.25  \\
 & & 3.00  & 12.00  & 0.00  & 0.00  & 0.00  & 0.00  & 1.00  & 35.00 \\
 \midrule 
  \multirow{12}{*}{HellaSwag} & \multirow{2}{*}{Translator} & \cellcolor{yellow!50}62.00   & \cellcolor{yellow!50}55.00   & \cellcolor{yellow!50}19.00   & \cellcolor{yellow!50}19.00   & \cellcolor{yellow!50}6.00   & \cellcolor{yellow!50}29.00   & \cellcolor{yellow!50}28.00   & \cellcolor{yellow!50}18.00  \\
 & & 51.00  & 22.00  & 0.00  & 0.00  & 0.00  & 0.00  & 24.00  & 7.00 \\
 & \multirow{2}{*}{Calculator} & \cellcolor{yellow!50}60.00   & \cellcolor{yellow!50}64.00   & \cellcolor{yellow!50}21.00   & \cellcolor{yellow!50}16.00   & \cellcolor{yellow!50}7.00   & \cellcolor{yellow!50}16.00   & \cellcolor{yellow!50}3.0   & \cellcolor{yellow!50}25.00  \\
 & & 52.00  & 22.00  & 1.00  & 0.00  & 0.00  & 3.00  & 25.00  & 7.00 \\
 & \multirow{2}{*}{Search Engine} & \cellcolor{yellow!50}40.00   & \cellcolor{yellow!50}50.00   & \cellcolor{yellow!50}21.00   & \cellcolor{yellow!50}10.00   & \cellcolor{yellow!50}4.00   & \cellcolor{yellow!50}27.00   & \cellcolor{yellow!50}0.00   & \cellcolor{yellow!50}29.00  \\
 & & 59.00  & 23.00  & 1.00  & 0.00  & 0.00  & 0.00  & 27.00  & 7.00 \\
 & \multirow{2}{*}{WikiPedia Search}& \cellcolor{yellow!50}41.00   & \cellcolor{yellow!50}31.00   & \cellcolor{yellow!50}19.00   & \cellcolor{yellow!50}18.00   & \cellcolor{yellow!50}0.00   & \cellcolor{yellow!50}21.00   & \cellcolor{yellow!50}31.00   & \cellcolor{yellow!50}21.00  \\
 & & 39.00  & 26.00  & 0.00  & 0.00  & 0.00  & 0.00  & 23.00  & 15.00 \\
 & \multirow{2}{*}{All} & \cellcolor{yellow!50}50.75   & \cellcolor{yellow!50}50.00   & \cellcolor{yellow!50}20.00   & \cellcolor{yellow!50}15.75   & \cellcolor{yellow!50}4.25   & \cellcolor{yellow!50}23.25   & \cellcolor{yellow!50}15.50   & \cellcolor{yellow!50}23.25  \\
 & & 23.00  & 28.00  & 1.00  & 0.00  & 0.00  & 0.00  & 4.00  & 11.00 \\
\bottomrule
     
\end{tabular}}
\caption{Detailed results of all general datasets experiment, where \textit{T003} means \textit{Text-Davinci-003}, and "\colorbox{yellow!50}{\quad}" indicates few-shot results, while cells without background color indicate zero-shot results.}
\end{table}

\twocolumn

\end{document}